\documentclass[acmlarge]{acmart}
\graphicspath{{./images/}}

\AtBeginDocument{%
  \providecommand\BibTeX{{%
    \normalfont B\kern-0.5em{\scshape i\kern-0.25em b}\kern-0.8em\TeX}}}

\setcopyright{acmcopyright}
\copyrightyear{2024}
\acmYear{2024}
\acmDOI{1111111.1111111}

\acmJournal{IMWUT}
\acmVolume{0}
\acmNumber{0}
\acmArticle{000}
\acmMonth{0}

\newcommand{\changed}[1]{\textcolor{black}{#1}}

\usepackage{booktabs}
\usepackage{graphicx}
\usepackage{subfigure}
\usepackage{textcomp}
\usepackage{geometry}
\usepackage{todonotes}
\usepackage{multirow}
\usepackage{comment}
\usepackage{url}
\usepackage{wrapfig}
\usepackage{wrapft}
\usepackage{mathtools}
\usepackage{graphicx}
\usepackage{multirow}
\usepackage{colortbl}
\usepackage{enumitem}
\usepackage{algorithm}
\usepackage{algorithmic}
\usepackage{xcolor}

\newcommand{\ourmethod}{\textit{EyeTrAES }}
\newcommand{\ourmethodf}{\textit{EyeTrAES's }}
\newcommand{\ourmethodnospace}{\textit{EyeTrAES}}

\newcolumntype{F}[1]{%
    >{\raggedright\arraybackslash\hspace{0pt}}p{#1}}%
\newcolumntype{T}[1]{%
    >{\centering\arraybackslash\hspace{0pt}}p{#1}}%

\definecolor{green}{HTML}{008B72}
\begin{document}


\title{\ourmethodnospace: Fine-grained, Low-Latency \underline{Eye} \underline{Tr}acking via \underline{A}daptive \underline{E}vent \underline{S}licing}

\author{Argha Sen}
\email{arghasen10@gmail.com}
\orcid{0000-0002-1579-8989}
\affiliation{%
  \institution{Indian Institute of Technology Kharagpur}
  \city{Kharagpur}
  \country{India}
}

\author{Nuwan Bandara}
\orcid{0009-0002-2509-5117}
\affiliation{%
  \institution{Singapore Management University}
  \country{Singapore}}
\email{pmnsbandara@smu.edu.sg}

\author{Ila Gokarn}
\orcid{0009-0001-6977-945X}
\affiliation{%
  \institution{Singapore Management University}
  \country{Singapore}}
\email{ingokarn.2019@phdcs.smu.edu.sg}

\author{Thivya Kandappu}
\orcid{0000-0002-4279-2830}
\affiliation{%
  \institution{Singapore Management University}
  \country{Singapore}}
\email{thivyak@smu.edu.sg}

\author{Archan Misra}
\orcid{0000-0003-1212-1769}
\affiliation{%
  \institution{Singapore Management University}
  \country{Singapore}}
\email{archanm@smu.edu.sg}


 


\begin{abstract}
\changed{Eye-tracking technology has gained significant attention in recent years due to its wide range of applications in human-computer interaction, virtual and augmented reality, and wearable health. Traditional RGB camera-based eye-tracking systems often struggle with poor temporal resolution and computational constraints, limiting their effectiveness in capturing rapid eye movements. To address these limitations, we propose \ourmethodnospace, a novel approach using neuromorphic event cameras for high-fidelity tracking of natural pupillary movement that shows significant kinematic variance. One of \ourmethodf highlights is the use of a novel adaptive windowing/slicing algorithm that ensures just the right amount of descriptive asynchronous event data accumulation within an event frame, across a wide range of eye movement patterns. \ourmethod then applies lightweight image processing functions over accumulated event frames from just a single eye to perform pupil segmentation and tracking (as opposed to gaze-based techniques that require simultaneous tracking of both eyes). We show that these two techniques boost pupil tracking fidelity by 6+\% 
, achieving IoU$\sim$=92\%, while incurring at least 3x lower latency than competing pure event-based eye tracking alternatives~\cite{li2024gaze}. We additionally demonstrate that the microscopic pupillary motion captured by \ourmethod exhibits distinctive variations across individuals and can thus serve as a biometric fingerprint. For robust user authentication, we train a lightweight per-user Random Forest classifier using a novel feature vector of short-term pupillary kinematics, comprising a sliding window of pupil (location, velocity, acceleration) triples. Experimental studies with two different datasets (capturing eye movement across a range of environmental contexts) demonstrate that the \ourmethodnospace-based authentication technique can simultaneously achieve high authentication accuracy ($\sim$=0.82) and low processing latency ($\sim$=12ms), 
and significantly outperform multiple state-of-the-art competitive baselines.}
\end{abstract}




\maketitle

\section{Introduction}
\label{sec:intro}

\changed{Fine-grained eye-tracking has become an increasingly important enabler of a variety of applications, spanning areas such as human-computer (gaze-based) interaction, consumer visual attention analysis, visuo-motor disease prediction~\cite{vodrahallibiocomputing2022}, Autism spectrum disorder identification~\cite{asmethajayarani2023}, and biometric authentication~\cite{lohr2022eyetoo, 9179933, komogortsev2010biometric}. Traditional eye-tracking systems typically rely on RGB cameras to capture images of the eye (often both eyes), which are then processed to extract information related to features such as gaze direction, fixation, and saccades. However, these systems often face challenges such as poor temporal resolution, relatively low frame rates, limited dynamic range, and high computational overheads. This is especially important as the eye muscles can generate powerful, short-lived but high-velocity movements. More specifically, the human eye is characterized by angular velocity exceeding speeds of $300^\circ/s$, particularly during saccadic eye motions~\cite{angelopoulos2021event}, and ocular acceleration reaching values as high as $24,000^\circ/s^2$~\cite{abrams1989speed}.}

\begin{figure}
    \centering
    \includegraphics[width=\textwidth]{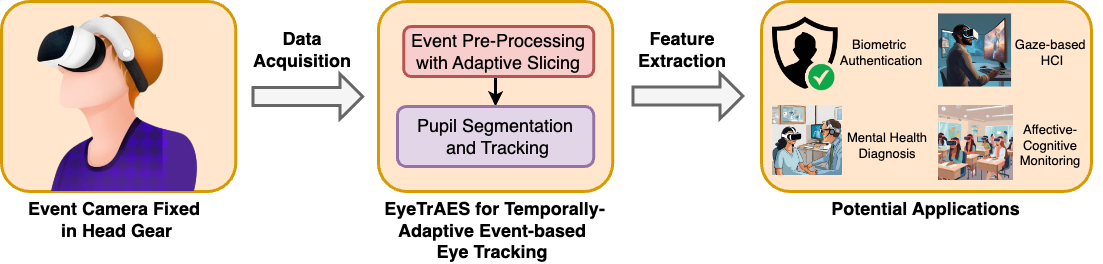}
    \caption{Broad Overview of \ourmethod{} 
    }
    \label{fig:overview}
\end{figure}

In recent years, neuromorphic event cameras have emerged as a promising alternative to traditional RGB cameras for eye-tracking applications~\cite{stoffregen2022event,angelopoulos2021event,zhao2024ev,li2024gaze,chen20233et}. Unlike traditional RGB cameras that capture frames at a fixed rate, event cameras produce events asynchronously, on a per-pixel basis, whenever there is a change in the incident intensity and with very low (O($\mu$sec)) latency. As a consequence, such cameras can not only operate at sampling rates of 10,000 Hz and above, but also result in their event generation rate adapting to the underlying event (i.e., eye movement) velocity. These unique characteristics make event cameras well-suited for capturing fast and dynamic phenomena, such as \emph{fine-grained} eye movements, with high temporal resolution and low latency.

\changed{A key challenge of such event camera-based eye tracking, besides the well-known lack of underlying color or texture information, is its extremely high event rate (as high as 1.06 Geps (Giga events per second) for dynamic scenes) and the consequent computational burden of event processing. To tackle these challenges, past work on ocular event data processing (e.g.,~\cite{li2024gaze}) first aggregates a collection of asynchronous events to create synchronous \emph{event frames}. Most eye tracking-based techniques first attempt to track the location (in the camera frame coordinates) of the eye pupil region within each such \emph{event frame} using standard image processing methods, and then extract pupillary movement-related features from a sequence of such `pupil location' values. In all past work, frame aggregation is performed over \emph{fixed windows} of either time or events. We shall, however, demonstrate that such a fixed windowing strategy is inadequate to support the wide \emph{variation} in eye movement rates (up to 700$^\circ$/sec during saccadic eye movements, as found in our studies): a small window captures insufficient kinematic data during periods of slow eye movement, whereas a large window is susceptible to \emph{over-accumulation}, and consequent loss of fine-grained movement information, during periods of rapid eye movement.}

\changed{In this paper, we explore the challenge of developing an event camera-based \emph{eye movement tracking} technique that has significantly higher spatiotemporal accuracy and is computationally efficient to support low-latency on-device execution. Our vision is that such miniaturized inward-facing event cameras can be mounted on personal wearable devices, such as augmented reality (AR) smartglasses or head-mounted displays (HMDs). By continuously tracking the wearer’s natural eye movement, such event cameras can support real-time extraction of pupillary kinematics. Based on results in prior physiological research, we believe that such accurate extraction of temporal pupillary movement features even under high kinematic variance can further increase the fidelity and variety of eye-tracking-based applications -- for example, rapid and intricate movements of eyes have been used in (a) diagnosing neurodevelopmental disorders (such as, ASD~\cite{asmethajayarani2023}, and ADHD~\cite{lee2023use}), (b) detecting low vision conditions~\cite{grootjen2023assessing}, (c) measuring cognitive load~\cite{duchowski2018index}, and (d) emotion recognition~\cite{zhang2023blink}.}

\changed{We shall show that the lack of adaptation in the accumulation window of current approaches consequently leads to a significant increase in pupil localization error, which in turn can lead to inaccurate computation of fine-grained temporal pupillary movement features. Our approach to on-device pupillary kinematic tracking using event cameras, called \ourmethodnospace\footnote{Eye Tracking via \underline{A}daptive \underline{E}vent \underline{S}licing, pronounced Eye-Trace} first proposes a novel \emph{adaptive slicing technique} to convert asynchronous event streams into event frames that capture just the right amount of pupillary kinematic information. We shall then present a new, classical image processing pipeline (in contrast to in-vogue heavyweight DNN based approaches) that can both extract the pupillary location coordinates from such event frames with high accuracy and is computationally lightweight enough to permit \emph{on-device} execution. We shall experimentally quantify the resulting improvements in real-time, on-device eye movement tracking and pupil localization. To additionally demonstrate the practical benefits of \ourmethodf improved pupillary tracking, we shall finally introduce a new biometric user authentication technique that utilizes a machine learning-based model, trained on natural pupillary \emph{micro-movement} features curated from individual-specific event-based eye movement data. Figure~\ref{fig:overview} depicts the sequence of novel capabilities that we shall demonstrate.} 

\noindent \textbf{Key Scientific Contributions:} 
\begin{itemize}[leftmargin=*]
\item [(a)] \emph{Develop an Adaptive Event Windowing and Time Slicing Technique:} We first demonstrate that past approaches for frame-based processing of event data (e.g.,~\cite{li2024gaze}), based on fixed windows of either time or events, are ill-suited to support the wide variation in eye movement rates (up to 700$^\circ$/s during saccadic eye movements, as found in our studies). Instead, \ourmethod employs a novel adaptive slicing technique to convert asynchronous event streams into event frames that capture just the right amount of pupillary kinematic information. Our approach maintains a running average of the (mean, standard deviation) of polarities of incoming events and demarcates a frame boundary only when the standard deviation exceeds a designated threshold. We shall show that this adaptive slicing approach results in a significantly lower error in pupil segmentation (30+\% improvement in IoU scores), across multiple segmentation approaches, compared to a conventional fixed window approach. 

\item [(b)] \emph{Ensure Low-Complexity, Low-Latency Pupil Tracking:} We develop an accurate and low-latency pupil tracking technique, amenable to \emph{on-device} execution, that pipelines a sequence of classical image processing techniques (such as edge detection, Hough transform based circle detection and Kalman filtering) over successive event frames. We show that \ourmethodf pupil segmentation module is superior in its ability to provide both accurate and low-latency pupil segmentation, achieving an IoU of 92\% and incurring a per-frame latency of 4.7 ms on a commodity Intel i9 workstation, compared to other alternatives such as Ev-Eye~\cite{zhao2024ev} (IoU=89\%, latency=700 ms), RGB based segmentation (IoU=84\%, latency=8 ms) and E-Gaze~\cite{li2024gaze}(IoU=87\%, latency=12 ms). 

\item [(c)] \emph{Establish the Distinctiveness of new Fine-grained (microscopic) Reflexive Eye Movement Features:} Drawing inspiration from studies in vision science and tracking~\cite{bargary2017individual, kasprowski2004eye, friedman2017method, komogortsev2010biometric, makowski2021deepeyedentificationlive, jia2018biometric}, we hypothesize that individual differences in ocular muscle strength generate distinct individual-specific natural spatiotemporal microscopic eye movement patterns. More specifically, we show that novel, short-duration kinematic features, such as \emph{pupillary velocity and acceleration}, together with the pupil’s location, effectively serve as a biometric fingerprint. Compared to prior approaches based on gaze features, \ourmethodf classifier requires tracking of only one eye and does not require calibration based on the gaze distance (distance to the viewed object). 

\item [(d)] \emph{Demonstrate \ourmethodf use in Biometric User Authentication:} Using data from both a publicly available prior dataset Ev-Eye~\cite{zhao2024ev} (10 users, but with fixed eye-screen distances) and an \ourmethod dataset collected via our own user studies (40 participants, no constraint on the eye-screen distance or on head movement), we shall demonstrate the superiority of user authentication using \ourmethodnospace-derived microscopic eye movement features. Our approach achieves significantly higher accuracy (median accuracy $\sim$0.82 on both datasets) than both (i) Ev-Eye (median accuracy of $0.62$ and $0.375$ for Ev-Eye and \ourmethod datasets, respectively), and (ii) a high frame rate (120 FPS) RGB camera-based authentication technique (median accuracy = $0.71$). Moreover, \ourmethod is able to perform such authentication rapidly, with an average authentication response time of $\sim$0.14 sec, depending on the kinematics of pupil movement.
\end{itemize}

\changed{Overall, we believe that \ourmethod dramatically improves the capability for pure event sensor-based pupillary movement tracking and resulting pupil movement-based user authentication, offering a compellingly superior alternative to extant \emph{gaze-based} methods that require significantly greater instrumentation and calibration. Furthermore, we have open-sourced our implementation for the research community, with the code and dataset accessible at https://anonymous.4open.science/r/EyeTrAES.}

\section{Related Work}\label{sec:relwork}
Eye-tracking technology has been the subject of extensive research and development in recent years, leading to a wide range of approaches and techniques for analyzing eye movements and extracting features for various applications. In this section, we review some of the key works related to eye-tracking using both RGB and event cameras, image processing methods for pupil tracking, and user authentication based on eye movements.

\textbf{RGB Frame-based Eye/Gaze Tracking}
Pupil tracking is a critical step in eye-tracking systems, as it provides information about the user's gaze direction and fixation points. To this end, most existing works for eye or pupil tracking are RGB frame-based approaches that utilize either traditional image processing methods~\cite{fuhl2016pupil, kassner2014pupil, dierkes2019fast, fuhl2015excuse, fuhl2016else, tonsen2017invisibleeye} or end-to-end deep learning methods~\cite{kim2019nvgaze, chaudhary2019ritnet, kothari2021ellseg, feng2022real}. 

With regard to the traditional image processing approaches for pupil tracking, methods including color filtering, ellipse fitting, and contour analysis are typically implemented in the literature such as~\cite{fuhl2016pupil}, in which the authors presented a method for pupil tracking using adaptive thresholding and ellipse fitting, achieving high accuracy in tracking pupil movements. Similarly, \cite{dierkes2019fast} utilized canny edge detection and blob detection to segment the frames and ellipse fitting to extract pupil features from the detected blobs while~\cite{kassner2014pupil} proposed a method for pupil detection using a combination of color segmentation and edge detection, demonstrating robust performance in various lighting conditions. To further build on these approaches and adapt them to real-world non-ideal scenarios, \cite{fuhl2015excuse} proposed a pupil detection method based on edge filtering and oriented histograms while~\cite{fuhl2016else} applied morphological operations and ellipse selection to build an eye-tracking system. However, with the rapid advancement in deep learning, most recent works tend to leverage neural networks for the task such as~\cite{chaudhary2019ritnet} in which the authors implemented a U-Net-based convolutional neural network architecture to segment the near-eye frame into four regions: background, sclera, iris and pupil. \cite{kothari2021ellseg} proposed an add-on regression module based on~\cite{chaudhary2019ritnet} to extract eye features from segmentation in order to fit ellipses for pupil and iris. 

\textbf{Event-based Eye/Gaze Tracking}
Event cameras~\cite{lichtsteiner2008128} have gained attention in the field of eye tracking due to their high temporal resolution, low latency, high dynamic range, sparse data acquisition, and asynchronous operation which eventually lead to capture rapid eye movements, such as saccades and microsaccades, with high precision and minimal motion blur, even in dynamic environments. 

Early works on eye or gaze tracking based on event cameras were predominantly proposed to detect faces or eye blinks via recording the subject's face, upper body or whole body~\cite{lenz2020event, ryan2021real}. Therefore, these setups were neither near-eye nor captured eye features or gaze features. Even though several works proposed to utilize near-eye setups to track eye and gaze starting from~\cite{angelopoulos2021event}, most of these works relied upon RGB frames as well since their proposed pipelines need to take both frame and event data as inputs. In~\cite{angelopoulos2021event}, the authors implemented a frame-based eye modelling pipeline and subsequently the captured event data was combined to update the eye model parameters. Eventually, the gaze direction was derived through a polynomial regressor. Following this work, numerous studies attempted to utilize both frame and event data for eye or gaze tracking such as~\cite{feng2022real, zhao2024ev, li2024denoising, zhang2024swift}. In~\cite{feng2022real}, collected event data was utilized to predict regions of interest and U-Net-like architecture was then implemented to perform eye segmentation whereas in~\cite{li2024denoising}, the authors suggested to utilize stacked event frames along with RGB frames in parallel, to predict the gaze location via a quantification network of state transitions. \cite{zhao2024ev} utilized a U-Net-based eye segmentation pipeline on the collected frames and then the binarized mask corresponding to the pupil area along with the event data were fed into a template-based pupil tracking stage. The prediction for the point of gaze was subsequently derived through a polynomial regressor. \cite{zhang2024swift} was proposed to address the issue of occlusion in eye tracking studies which first interpolated RGB frames with the event data and then utilized a deep multi-scale spatial extraction-fusion network and an anti-blink pupil estimation module to extract semantic information from different scales and to deal with involuntary blinks respectively. However, due to the dependency on RGB frames, these works are unable to fully harness the benefits of the sparse, asynchronous and low-power characteristics of event cameras and thus, the need for an event camera-exclusive eye tracking system is still not properly addressed.

To this end, several recent works~\cite{stoffregen2022event,  li2023track, chen20233et,  bonazzi2024retina, li2024gaze, wang2024event} attempted to develop fully event-based eye tracking systems. In~\cite{stoffregen2022event}, even though the authors proposed an event-based eye tracking system without being dependant on RGB frames, their system heavily relied upon LEDs to generate glints as markers to execute corneal sphere regression and further the implemented coded differential lighting scheme on LEDs was limited by the sensor bias of the event cameras, especially when operating at high frequency. \cite{li2023track} proposed an convolutional neural network-based pupil tracking system operating on event frames while~\cite{chen20233et} proposed to replace convolutional neural network architecture with a change-based convolutional long short-term memory network for better performance. \cite{bonazzi2024retina} further attempted to reduce the model complexity via implementing a spiking neural network and claimed better precision in localizing the pupils than~\cite{chen20233et} with fewer computational complexity. More recently, \cite{li2024gaze} proposed to utilize traditional kernel and ellipse fitting methods on event frames, which were accumulated over a fixed number of events, to extract pupil features and subsequently the pupil feature vector was fed into a recurrent neural network to predict the gaze location. However, all these methods still lack the operational capability to run real-time or near-real-time to predict pupil location using event data. 


\textbf{Event Accumulation}
Most works in the literature follow two approaches when it comes to event accumulation: a fixed time interval~\cite{zhang2024swift, lagorce2016hots} or a fixed number of events~\cite{angelopoulos2021event, zhao2024ev, li2024gaze}. In the context of a fixed time interval, the scene dynamics, here the eye motion, may lead to fewer (if the eye moves slightly or does not move at all) or higher (if the eye moves fast) number of events in the accumulation process which eventually result in poor performance in downstream task due to scene instability occurred within a same time interval. On the contrary, the utilization of a fixed number of events seems better since the events are driven by the motion and thus a fixed number of events represents a consistent and stable amount of motion. Further, unlike the fixed time interval approach, the later approach also preserves the asynchronous nature of events by allowing the accumulation to be executed asynchronously. However, optimally determining the fixed number for later approach is challenging: if the selected threshold is too large, the accumulated events will present motion blur where as the threshold is too low, the accumulated events will lack the motion information~\cite{zhang2022adaptive}. In addition, a pure event count based approach does not consider the \emph{informativeness} of the underlying events--e.g., whether they are generating by motion of multiple eye segments vs. motion of a single segment. Therefore, there is a need for an adaptive technique for event accumulation (or slicing) technique which is based on the spatiotemporal dynamics of the scene rather than on an artificially determined threshold.

The use of adaptive downsampling and accumulation for processing event data has been explored in few works. In \cite{nunes2023adaptive} authors proposed an adaptive event camera that adjusts the event rate based on the scene's motion characteristics, leading to improved tracking performance in dynamic environments. Similarly, \cite{zhang2022adaptive} introduced a method for adaptively accumulating events based on their relevance to the scene, reducing the data rate while maintaining important information for tracking tasks. These works demonstrate the effectiveness of adaptive downsampling and accumulation in improving the interpretability and efficiency of event data processing.

\textbf{Eye Movements-based User Authentication}
User authentication based on eye movements has been explored as a biometric authentication method in numerous studies~\cite{kasprowski2004eye, friedman2017method, komogortsev2010biometric,makowski2021deepeyedentificationlive, jia2018biometric, lohr2022eyetoo, zhang2018continuous, khamis2018cueauth}. Starting from~\cite{kasprowski2004eye} in which the eye movement biometric modality was introduced, earlier works such as~\cite{friedman2017method} required to explicitly classify the eye motion signals into physiologically-grounded events and subsequently the manually-extracted features were fed into statistical models. With the advancement of deep learning, several works were proposed as end-to-end pipelines for eye movement biometrics. \cite{jia2018biometric} presented a task-independent recurrent neural network-based architecture for human identification using gaze points whereas~\cite{makowski2021deepeyedentificationlive} utilized two convolutional subnets to separately focus on saccadic and fixational gaze movements. More recently, \cite{lohr2022eyetoo} claimed to acquire higher performance by utilizing a parameter-efficient DenseNet-based~\cite{huang2018densely} deep architecture. However, all these works are explicitly dependant on the gaze estimations from an off-the-shelf eye tracker and therefore, neglect the potential of utilizing the motion dynamics of the pupil as a promising eye movement biometric.    

To this end, only few works exist in the literature, such as~\cite{9179933, komogortsev2010biometric}, in which the movement of pupil is utilized to derive the authentication features. \cite{9179933} proposed a pupil movement-based system for personal identification number generation in which a set of classifiers were set to identify face, eye, eye-blinks and pupil movement. However this work is highly task-dependent and does not fully harness the motion dynamics of the pupil. In \cite{komogortsev2010biometric} authors investigated the use of eye movement characteristics, such as saccades and fixations, for user authentication and showed that these characteristics can serve as reliable biometric identifiers. They also highlighted the importance of capturing individual differences in eye movements for authentication purposes, which aligns with our motivation for using eye movements as biometric features. However, their system was still dependent on the gaze estimations from an eye tracker to extract the eye movement states in the eye movement classification block. 

\changed{Katsini et. al.~\cite{katsini2020role} surveyed the role of eye gaze in security and privacy applications. They discuss the evolution of eye tracking technologies and their application in biometric authentication, password entry, and privacy protection. Besides describing how eye tracking algorithms can be integrated into various devices such as smartphones and head-mounted displays, the authors also identify promising research directions and challenges for gaze-based security applications. More recently, Lien and Bhadhuri~\cite{lien2023challenges} explored the challenges and opportunities of biometric user authentication in the context of IoT devices. Their survey highlights the potential of biometric authentication to complement traditional knowledge-based methods. They categorize biometric traits into stable and volatile traits, a paradigm that aligns with our proposed approach of using natural eye kinematics as a stable biometric feature for enhanced, continuous and secondary authentication.}

In summary, the related work highlights the potential of event cameras for eye-tracking applications, the effectiveness of adaptive event accumulation for processing event data, and the importance of pupil tracking and eye movement analysis for user authentication. Our work builds upon these existing approaches by proposing a novel system for accurate temporal tracking of microscopic eye movement using event cameras, and subsequently using such eye kinematic features to support secure and efficient authentication.

\section{Background and Motivation}
\label{sec:background}

\changed{In this work, our primary objective is to capture reflexive physiological eye movements with high temporal resolution while also ensuring that downstream image processing methods operating on \emph{event frames} are able to track pupil movement with low computational complexity}. In this section, we provide a brief background on event cameras and our rationale behind various design choices of our system \ourmethod.

\subsection{2D Virtual Event-frame Construction}
Contrary to the conventional cameras (where the intensity of light across the visible spectrum incident on the sensor is captured at discrete points in time), event cameras or Dynamic Vision Sensors (DVS) only record changes in brightness (events) at each pixel asynchronously and with high temporal resolution, resulting in sparse data streams that encode motion and brightness changes in real-time. Event cameras are particularly beneficial in scenarios with either (i) low lighting/illuminance or (ii) high-speed motion where conventional cameras may suffer from motion blur and yield lower downstream task accuracy. In Section~\ref{sec:implementation}, we compare the impact of illuminance on the accuracy of downstream pupil movement-based user authentication \changed{(an exemplar application of our technique)} as observed from RGB and event cameras. We show that under low lighting conditions (environment illuminance of 24 lux and near-eye illuminance of 8 lux), RGB-based pupil detection methods suffer a steeper reduction in user authentication of $\sim45\%$ while event cameras suffer a moderate accuracy drop of $\sim33\%$ when compared to the achievable accuracy in standard lighting conditions for both cameras (environment illuminance of 348 lux and near-eye illuminance of 65 lux). While event cameras also suffer from a reduction in user authentication accuracy due to the reduced illumination, they are able to recover $\sim12\%$ of user authentication capability due to the generation of events even under low-lighting conditions, while RGB-based pupil detection and authentication relies on color-based filtering techniques which fail under poor lighting conditions. We also show in Sections ~\ref{sec:system} and ~\ref{sec:implementation} how \ourmethod leverages event cameras to adapt to different pupil kinematics patterns and high-speed pupil motion. 


The event camera outputs a series of events on a per-pixel level -- an event $e_i$ ($i \in \mathbb{N})$ is denoted by a tuple $(x_i, y_i, p_i, t_i)$, where $(x_i, y_i)$ denotes the corresponding pixel coordinates where the event is generated, $p_i$ represents the change in polarity (positive vs. negative), and $t_i$ is the time of the corresponding event. To efficiently process the sparse stream of spatiotemporal asynchronous events, traditionally, the events are accumulated periodically over a time interval $T$ or a fixed event volume $N$ to generate a virtual frame. Subsequently, the appropriate vision techniques tailored to the specific downstream task (e.g., object detection or tracking) are applied. However, such simple motion-oblivious event accumulation techniques pose several drawbacks especially when tracking dynamic scenes with rapid movements: (i) blurred motion representation (especially when more than an ideal number of events are aggregated), (ii) loss of motion dynamics (occurred when the accumulated events do not accurately capture motion dynamics, i.e., fewer than necessary events are accumulated), and (iii) high computational load (event cameras can generate over 1 Gigaevents per second of data).

\begin{figure}
    \centering
    \subfigure[]{
    \includegraphics[width=0.3\textwidth]{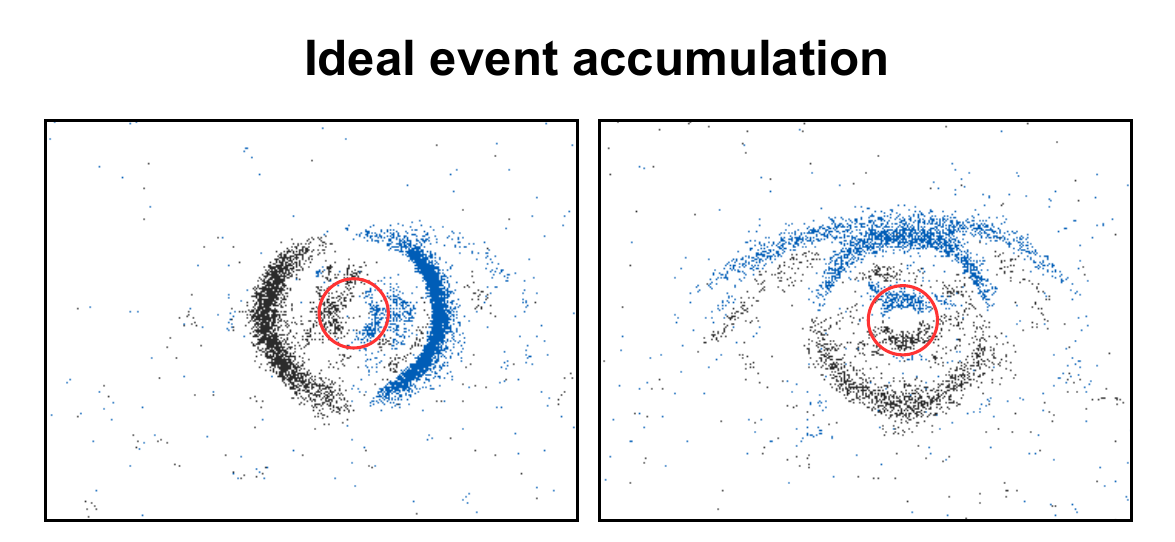}
    \label{subfig:right_slice}
    }
    \subfigure[]{
    \includegraphics[width=0.3\textwidth]{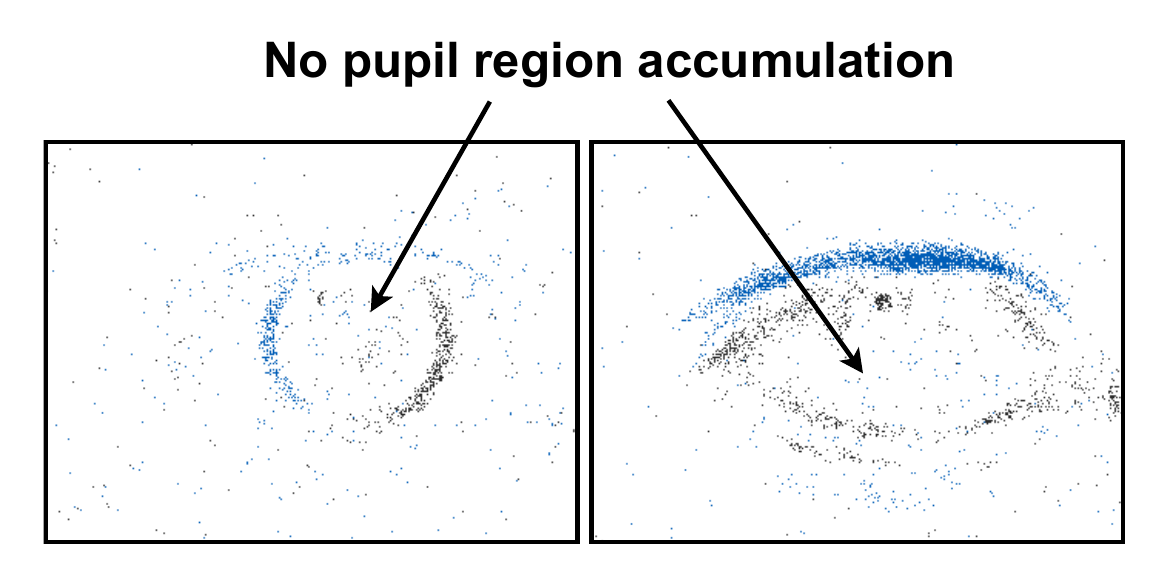}
    \label{subfig:small_slice}
    }
    \subfigure[]{
    \includegraphics[width=0.3\textwidth]{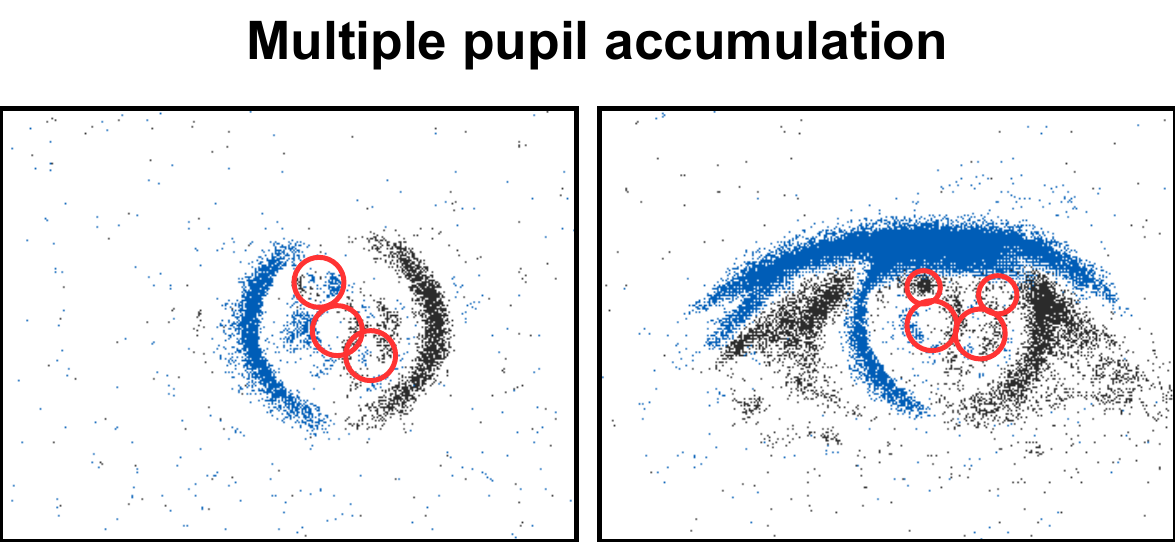}
    \label{subfig:large_slice}
    }
    \subfigure[]{
    \includegraphics[width=0.3\textwidth]{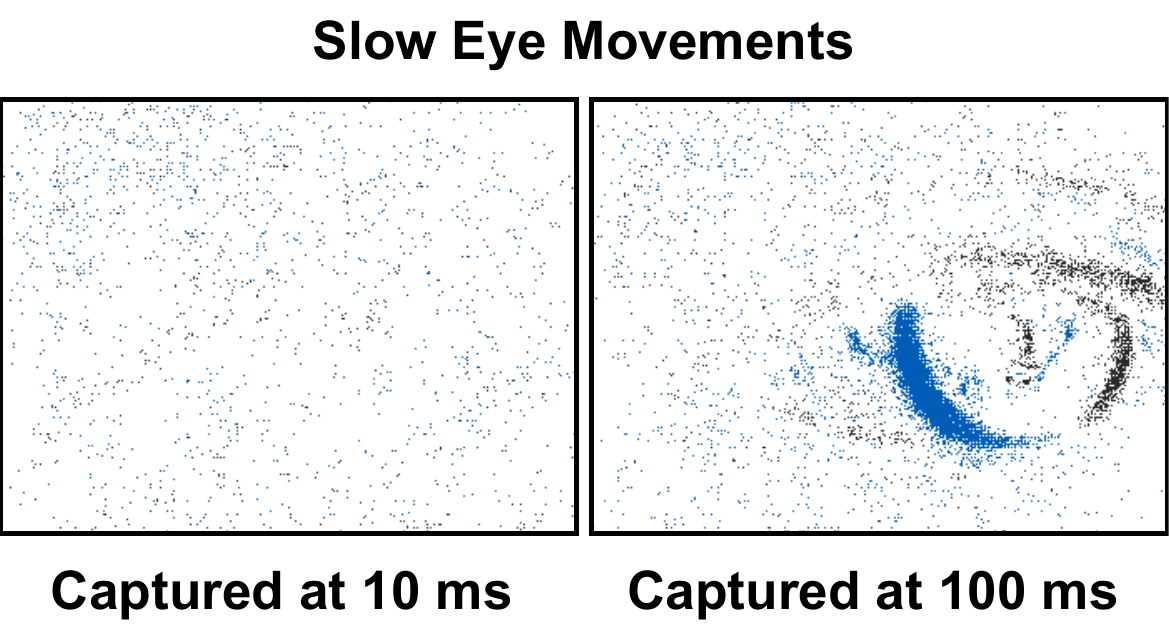}
    \label{subfig:slow}
    }
    \subfigure[]{
    \includegraphics[width=0.3\textwidth]{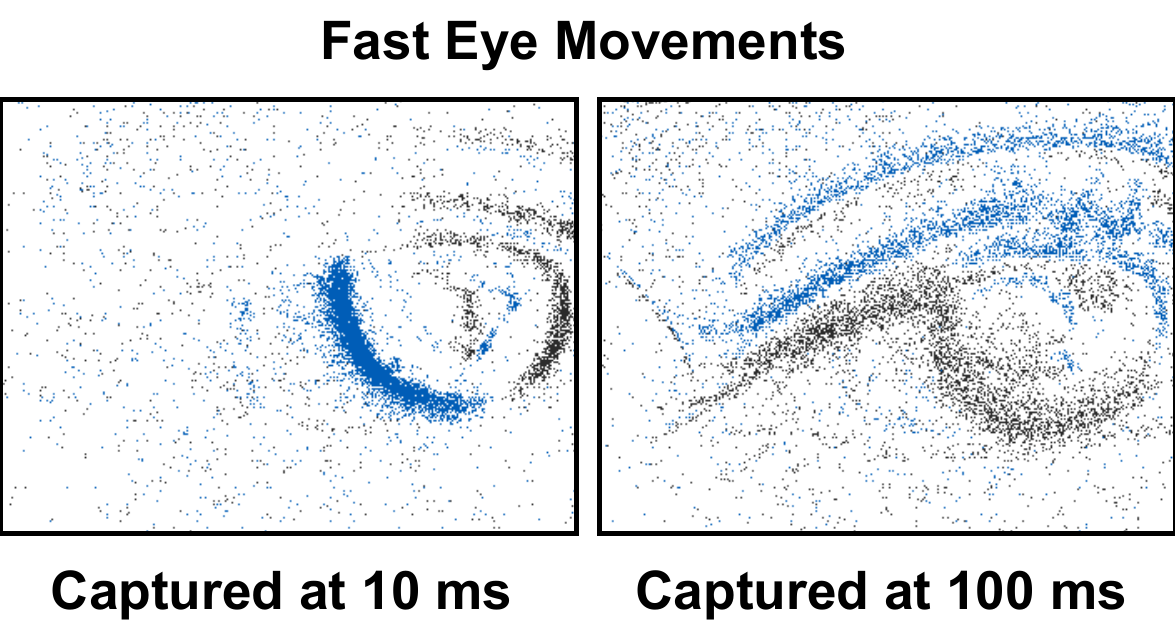}
    \label{subfig:fast}
    }
    \caption{Temporal Event Accumulation for both Normal and Eyelash-Flickering Eye Movement: (a) ideal event representation (duration = 30 ms), (b) Under-Accumulation (duration = 10 ms), (c) Over-Accumulation (duration =100 ms), (d) Slow eye movements captured at 10 and 100 ms, (e) Fast eye movements captured at 10 and 100 ms.}
    \label{fig:different_slice}
\end{figure}

To visualize the importance of appropriate event accumulation, in Figure~\ref{fig:different_slice}, we depict the 2D frame representation of the accumulated events over varying time windows for our targeted task of pupil tracking using a near-eye event camera. Each point in the image represents an event while the color of the point (blue vs. black) denotes the polarity. We illustrate two different pupil event frames, one for normal eye movement and the other including additional movement of the eyelashes during a blinking action. Figure~\ref{subfig:right_slice} denotes the ideal event accumulation, where a single pupil is detected in the 2D frame (duration= 30 ms). As shown in Figure~\ref{subfig:small_slice} and~\ref{subfig:large_slice}, accumulating fewer events (i.e., a frame with a lower duration= 10 ms) may not capture adequate motion artifacts, causing a failure to detect any pupil contours, whereas aggregating too many events (i.e., a frame with a larger duration= 100 ms) results in the detection of multiple pupils.  Importantly, the ideal frame duration is not constant, but varies with the speed and the spatial dynamics of the pupil movement.  

To further validate this point we explicitly take two scenarios where the subject is asked to have faster and slower eye movements in two different sessions. As shown in Figure~\ref{subfig:slow} and~\ref{subfig:fast}, during slower eye movements, a smaller accumulation window can't capture the eye movements at all, while for a rapid eye motions having a longer accumulation window can lead to noisy framed representations.  

This evidence raises the question: can the virtual 2D-frames be composed in a \emph{motion responsive} manner, such that the spatiotemporal event stream is adaptively ``sliced'' (or aggregated) based on the underlying rate and spatial dynamics of the event stream? In Section~\ref{subsec:adaptiveslicing}, we shall introduce the technique for such \emph{adaptive slicing}, as well as empirically demonstrate that natural variations in the speed of human pupil movement translate into significant variations in the ``optimal'' framing duration. We refer to ``event slices" as ``event frames" and ``framed representations" interchangeably in this work. 

\subsection{Morphological Segmentation for Eye/Gaze Tracking}

Conventional methods of RGB frame-based gaze tracking (captured using two RGB cameras) involve three steps: (i) morphological segmentation of the eye (i.e., segmenting eye parts using vision algorithms, such as canny edge detection), (ii) extraction of 2D eye features for tracking, and (iii) estimation of the direction of human gaze by mapping the 2D eye features extracted from \textit{both left and right eyes} using geometric approaches. 

Recent works on event-based gaze tracking involved event cameras that simultaneously capture \emph{both} asynchronous event streams and corresponding RGB frames; for example, the DAVIS346~\footnote{https://inivation.com/wp-content/uploads/2023/11/2023-11-iniVation-devices-Specifications.pdf} camera records frames at ~30 FPS. These works optimize the accuracy of gaze detection by using sparse RGB frames and initiate the pupil segmentation pipeline using vision algorithms to prime the event-based high-frequency pupil tracking (captured by two event cameras). Subsequently, a polynomial regressor is used to translate the features, representing pupillary information from both eyes, into 3D gaze direction.

We adopt an approach that differs from such prior work in the following ways: 
\begin{itemize}[leftmargin=*]
    \item \textbf{Pupil Tracking:} In our \ourmethod approach, we focus on tracking the pupil's spatial coordinates rather than the trajectory of the \emph{gaze direction}: pupil trajectory refers to the movement of the pupil within the eye over time, while the gaze direction represents the direction in which a person is looking relative to their environment. While correlated, pupil trajectory and gaze direction represent different aspects of eye movement. For example, changes in lighting conditions or cognitive load can cause fluctuations in pupil movement without necessarily corresponding to changes in gaze direction. Similarly, reflexive eye movements, such as saccades or smooth pursuit, can cause rapid changes in gaze direction while the pupil trajectory remains relatively stable.  
    Given our end goal of capturing fine-grained, microscopic and reflexive eye movements, we focus on pupil tracking in contrast to the dominant approach of gaze tracking. Additionally, we shall show (Section~\ref{subsec:authperf}) that we can achieve user authentication with higher accuracy and shorter observational period (often less than 120-200 ms) via the use of pupillary kinematic features, instead of gaze-based features.

    \item \textbf{Single Near-Eye Event Camera:} As we rely on tracking the reflexive, physiologically-driven spatiotemporal dynamics of the pupil, we require the use of only a single near-eye event camera. Our approach of tracking a single pupil is based on the assumption that users demonstrate ideal conjugate eye movements, reflecting synchronized ocular motions consistent with typical oculomotor function. Such single pupil tracking also allows us to reduce the computational complexity of the event processing pipeline. This choice differs from the conventional approaches, across both RGB and event-based methods, that employ dual-camera setups.

    \item \textbf{Exclusive Event-Based Pupil Tracking:} While most prior work utilizes both RGB frames and event streams for gaze tracking, a few recent works have focused on event-only data processing. E-gaze~\cite{li2024gaze} proposes an approach for purely event sensing-based gaze estimation based on the accumulation of event data into virtual \emph{event frames}. In E-Gaze, after segmenting the different parts of the eye, the proposed approach leverages kernel density to find the pupil center. We adopt a similar approach, using contour detection together with Kalman filtering to support reliable tracking of pupil location over consecutive frames. However, we shall demonstrate that our method is computationally cheaper, incurring $\sim$70\% lower latency compared to E-Gaze. More recently,  Retina~\cite{bonazzi2024retina} integrates a spiking neuron network (SNN) and a specialized hardware accelerator (SynSense Speck~\footnote{https://www.synsense.ai/products/speck-2/}) for energy-efficient eye tracking of the pupil from near-eye events. However, such hardware accelerators are not widely available, precluding practical execution of SNN pipelines on current resource-constrained wearable devices. 
    

\end{itemize}

\subsection{Eye Movements for Biometric Authentication}
Besides demonstrating an improvement in pupil micro-movement tracking, we shall also use eye-movement based user authentication as an exemplar to illustrate the application-level benefits of \ourmethodnospace. 
Our approach to biometric authentication is based on the assumption that the reflexive involuntary eye movements of different individuals, in response to simple naturally occurring visual stimuli, are distinctive and driven by natural physiological variations (e.g., in ocular muscle strength). Prior work, such as \cite{kasprowski2012first, george2016score}, has used statistical features extracted from the position, velocity and acceleration profiles of the \emph{gaze} sequence for applications such as task classification or user authentication. Individuals are assumed to exhibit distinctive patterns for gaze-related artifacts, such as saccades, fixations, and smooth pursuits. For \ourmethodnospace, we aim to use lower-cost \emph{pupil movement-based} proxies for artifacts such as fixations and saccades; more specifically, instead of attempting to explicitly identify such artifacts, we compute and utilize statistical features, such as the velocity and acceleration profiles for the pupil trajectory, of a single eye.

\section{\ourmethod Overview \& Functional Details}
\label{sec:system}

\begin{figure*}[thb]
        \centering
        \vspace{-0.1in}
        \includegraphics[width=0.8\textwidth]{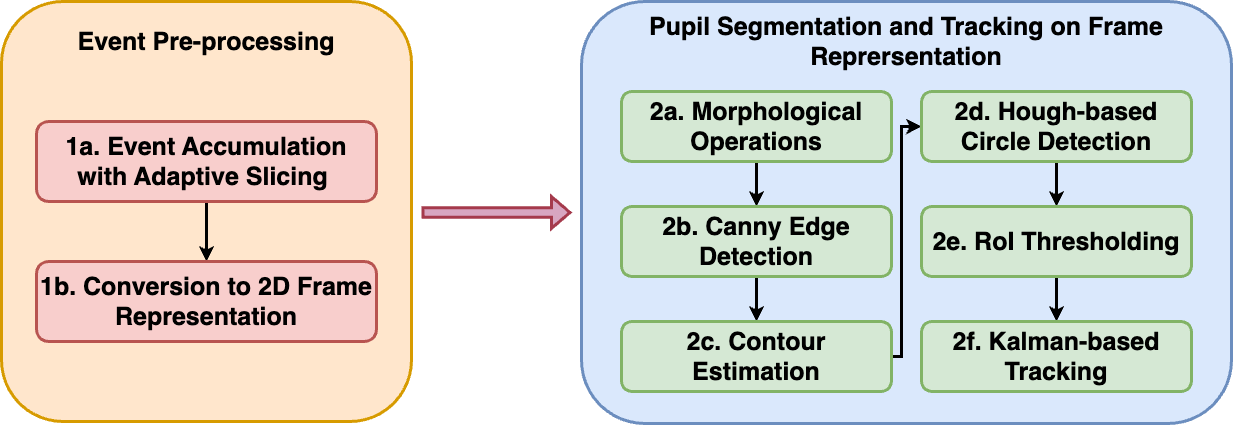}
        \caption{\ourmethodnospace: Block Diagram of Sub-Components}
        \label{fig:aes_big_picture}
        \vspace{-0.1in}
\end{figure*}

We now describe the core components (illustrated in Figure~\ref{fig:aes_big_picture}) in \ourmethodnospace: namely, the (i) event-based data acquisition and adaptive slicing that helps create appropriately informative 2D framed representations, and (ii) lightweight, signal processing based pupil segmentation and tracking that operates on a sequence of such 2D event frames.


At a high level, \ourmethod first ingests high-volume event data captured at $O(\mu sec)$ latency and determines the appropriate rate (or boundary) at which events are accumulated into a 2D framed representation. Such processing deviates from standard event accumulation techniques, which aggregate events either over a fixed/periodic time horizon or fixed event volume, and allows \ourmethod to autonomously adapt to the rate of change of information (using statistical measures) being captured by the event camera. \ourmethod then collates the events into a 2-channel frame (one for each polarity) for further processing. The rest of \ourmethodf Pupil Tracking pipeline then evaluates these 2D framed representations for (i) pupil segmentation using traditional OpenCV methods for canny edge detection~\cite{canny} and circle detection of the pupil using the Hough algorithm~\cite{hough}, and (ii) tracking of the segmented pupil using a Kalman-based Centroid tracker. 




\subsection{Events Preprocessing with Adaptive Slicing}
\label{subsec:adaptiveslicing}


We employ a neuromorphic event camera to capture asynchronous events representing changes in luminance. The event camera provides high temporal resolution and low latency, making it suitable for capturing rapid eye movements, such as saccades and microsaccades. The event data is streamed continuously and processed in real-time. In addition, the event camera also generates RGB frames at the nominal frame rate of 30 FPS.


\begin{algorithm}
\caption{Adaptive Event Slicing}\label{algo:adapt_slice}
\begin{algorithmic}[1]
\REQUIRE Continuous stream $E_{(x,y,p)}$, threshold $thresh$
\STATE Initialize running mean and standard deviation $\mu_{running} = 0$, $\sigma_{running\_prev} = 0$
\STATE Initialize Slice as an empty set
\STATE Initialize Frame as a matrix of size $H \times W$ filled with zeros
\STATE Set downsample factor $d = 2$
\FOR{each $(x, y, p)$ in $E$}
    \STATE Add $(x, y, p)$ to Slice
    \IF{$x \% d == 0$ or $y \% d == 0$}
        \STATE $p_f = |p|$
        \STATE $\text{Frame}[x, y] = p_f$
        \STATE $\mu_{current} = \text{mean}(F)$
        \STATE $\sigma_{current} = \sqrt{\frac{1}{H \times W} (\mu_{current} - p_f)^2}$
        \STATE $\sigma_{running} = (1 - \frac{1}{\text{Number of events}})\sigma_{running\_prev} + \frac{1}{\text{Number of events}}\sigma_{current}$
        \STATE $\sigma_{running\_prev} = \sigma_{running}$
        \IF{$\sigma_{running} > thresh$}
            \RETURN Slice
        \ENDIF
    \ENDIF
\ENDFOR
\end{algorithmic}
\end{algorithm}

\begin{figure}
    \centering
    \subfigure[]{
    \includegraphics[width=0.45\textwidth]{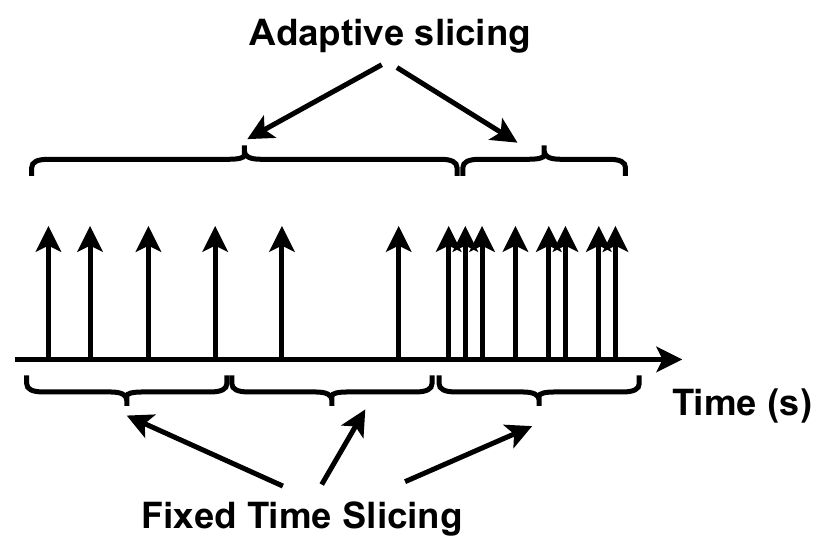}\label{fig:adaptive_slice_over}
    }    
    \subfigure[]{
    \includegraphics[width=0.45\textwidth]{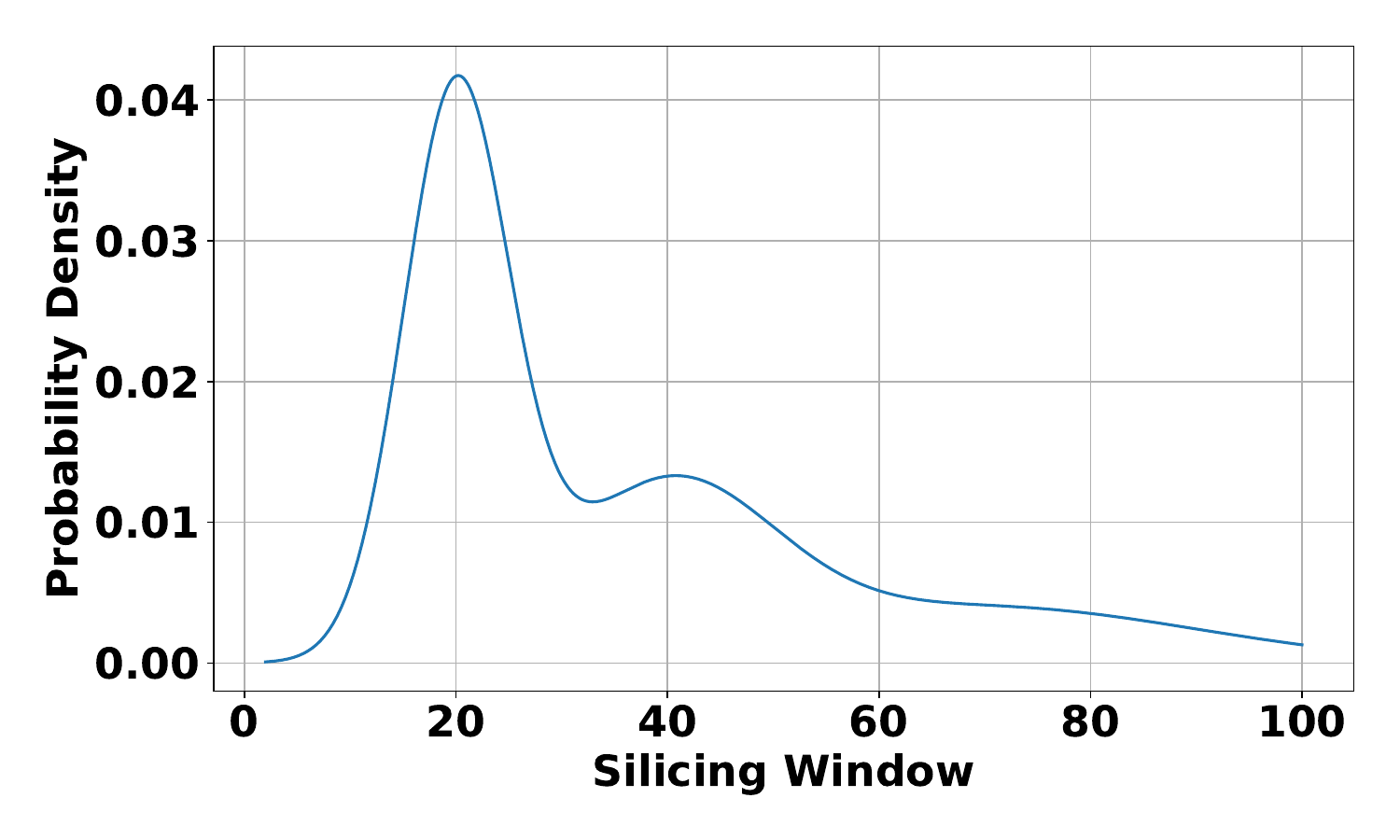}
    \label{fig:pdf_adaptive_distribution}
    }
    \caption{(a) Event splitting with adaptive slicing vs default fixed time slicing, and (b) Probability density across different slicing window length (in ms)}
    \label{fig:adaptiveSampling}
\end{figure}

The Event Pre-processing component of the \ourmethodf pipeline, illustrated in orange in Figure~\ref{fig:aes_big_picture}, accumulates the arriving asynchronous events to create \emph{event frames}. The adaptive slicing-based accumulation adjusts the event accumulation rate based on the content of the events stream, ensuring that an event frame is \emph{right sized} to capture just an adequate amount of data, as well as to reduce the overall frame rate. As summarised in Algorithm~\ref{algo:adapt_slice}, our method involves iterating through the continuous event stream and calculating (i) the sliding mean and (ii) the standard deviation for each pixel's polarity change. (These statistics are computed using the changes in the absolute values of each event's polarity attribute.)
If the standard deviation surpasses a predefined threshold $th$ (set to 0.001, based on empirical studies), we segment the event stream into a \emph{slice}; otherwise, we continue accumulating events. We also incorporate a downsampling factor of 2 to decrease the statistical mean and standard deviation computations, aiding in more efficient data processing while retaining crucial information. Our proposed slicing strategy leads to a variable slicing duration (i.e., a variable inter-frame gap): as illustrated in  Figure~\ref{fig:adaptive_slice_over}, the slice window is smaller in case of rapid eye movements (larger event volumes with higher polarity variations) and longer in case of slower eye movements (lower event volumes).


For visual reference, in Figure~\ref{fig:pdf_adaptive_distribution} we plot the probability density of the length of slices (in ms) in our adaptive slicing method using the publicly available Ev-Eye dataset~\cite{zhao2024ev} (described in detail in Section~\ref{subsec:ev-eye_dataset}). 
As we can see, during involuntary eye movements, both slower (smaller event volume) and rapid (larger slices) eye movements occur more organically, indicating that adopting a motion-agnostic fixed-time or fixed-event volume-based slicing techniques may fall short in achieving our overall goal of accurately capturing intricate spatiotemporal eye movement dynamics.

\begin{figure}
    \centering
    \includegraphics[width=0.7\textwidth,height=1.8in]{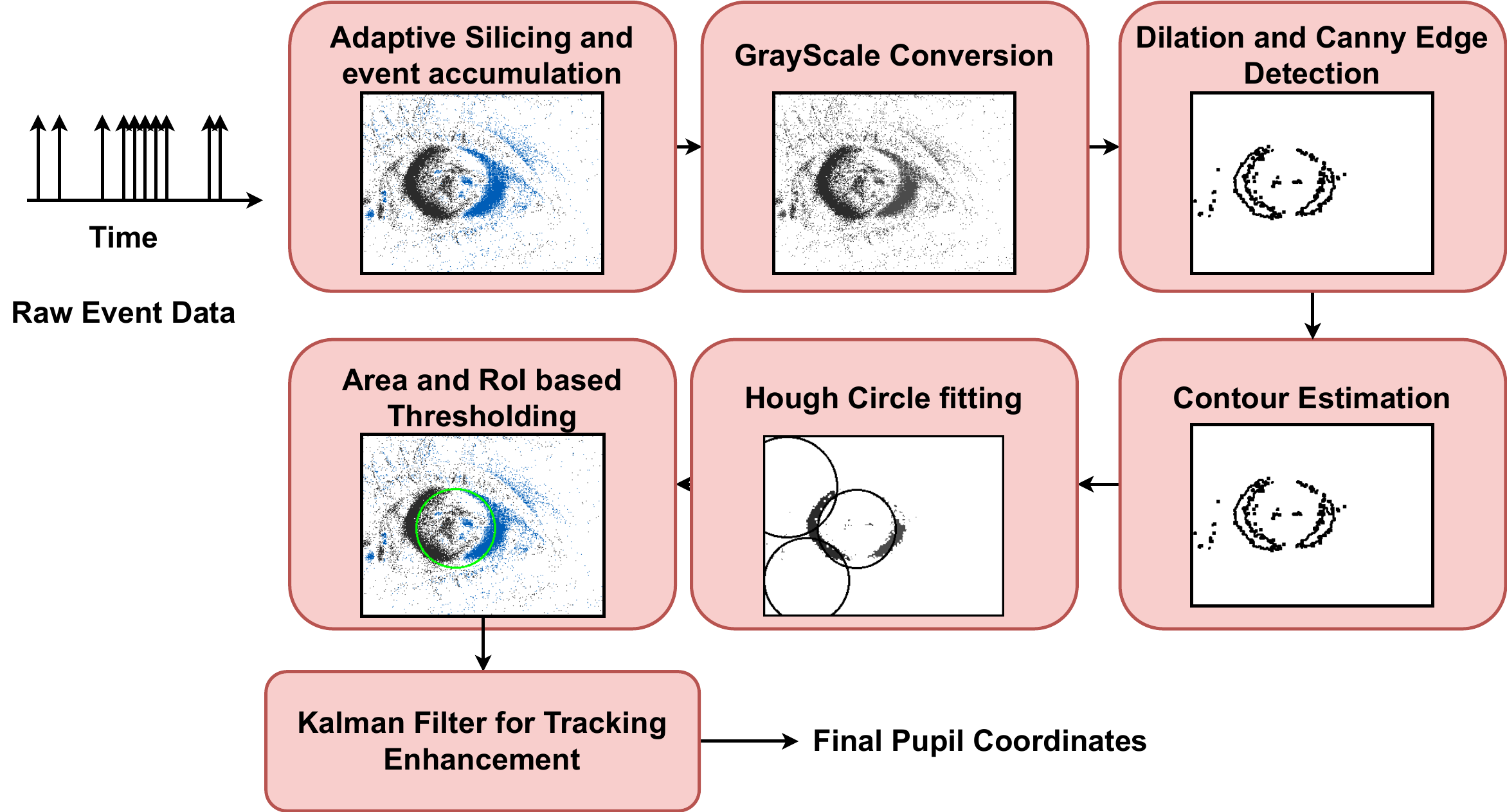}
    
    \caption{Event-based pupil segmentation}
    \label{fig:event_based}
\end{figure}

\subsection{Pupil Segmentation and Tracking}
\label{subsec:pupil_tracking}
Most prior work (e.g.,~\cite{zhao2024ev, angelopoulos2021event}) has utilized a frame-based pupil segmentation approach, with event data serving merely as a supplementary input to refine the segmented pupil region. However, when the initial template for the segmented pupil region is inaccurate, the addition of event information can introduce more noise, potentially degrading the overall pupil segmentation performance. In this work, we introduce a novel, fully event-based pupil segmentation approach (illustrated in blue in Figure~\ref{fig:aes_big_picture}) over the accumulated framed representation of events. As shown in Figure~\ref{fig:event_based}, the framed representation of the adaptively accumulated events primarily exhibits two distinct colors: blue, symbolizing positive polarities, and black, representing negative polarities. However, due to the inherent sparsity of data in the captured events, we utilize a sequence of preprocessing steps (described next) to enhance the visibility of relevant features. The output of each component in this sequence is also illustrated in Figure~\ref{fig:event_based}.

\noindent\textbf{Morphological Operations:} Initially, we convert the frame representation to grayscale to simplify subsequent processing steps. A dilation morphological operation is applied to expand the foreground event regions, facilitating better feature extraction.

\noindent\textbf{Canny Edge Detection:} Subsequently, the dilated frame undergoes canny edge detection~\cite{canny}, a widely used technique for identifying sharp transitions in pixel intensity, thereby delineating the boundaries of significant event contours. This step is crucial for isolating regions of interest corresponding to meaningful event occurrences within the frame. 

\noindent\textbf{Contour Estimation:} Further refinement is achieved through contour estimation, wherein continuous boundaries connecting adjacent pixels are identified by analyzing variations in pixel intensity. This process enables the extraction of precise contours outlining significant event regions, laying the groundwork for subsequent analysis.

\noindent\textbf{Hough-based Circle Detection:} We employ the classical Hough technique~\cite{hough} used for identifying and characterising complex shapes such as pupils accurately. This technique is particularly robust against gaps in curves and noise, making it well-suited for our application. By detecting circles that approximate the contours of interest, we can effectively estimate unknown boundaries and discern potential pupil regions within the frame.

\noindent\textbf{RoI Thresholding:} We next apply area and region-based thresholding techniques over the detected contours to isolate candidate elliptical shapes resembling human eye pupils. By imposing constraints on the size and shape of the detected regions, we enhance the precision of pupil localization (eliminating false positives), thus improving the overall accuracy of our system.

\noindent\textbf{Kalman-based Tracking:} We incorporate a Kalman filter to refine the pupil detection process further and mitigate the effects of noise. By recursively updating and refining the estimated pupil centres across successive frames, the Kalman filter helps denoise the detections and improve the stability of the overall tracking process. This adaptive filtering approach ensures robust and reliable pupil localization, even in the presence of varying lighting conditions and occlusions.
\begin{figure}[!h]
    \centering
    \subfigure[]{
    \includegraphics[width=0.45\textwidth]{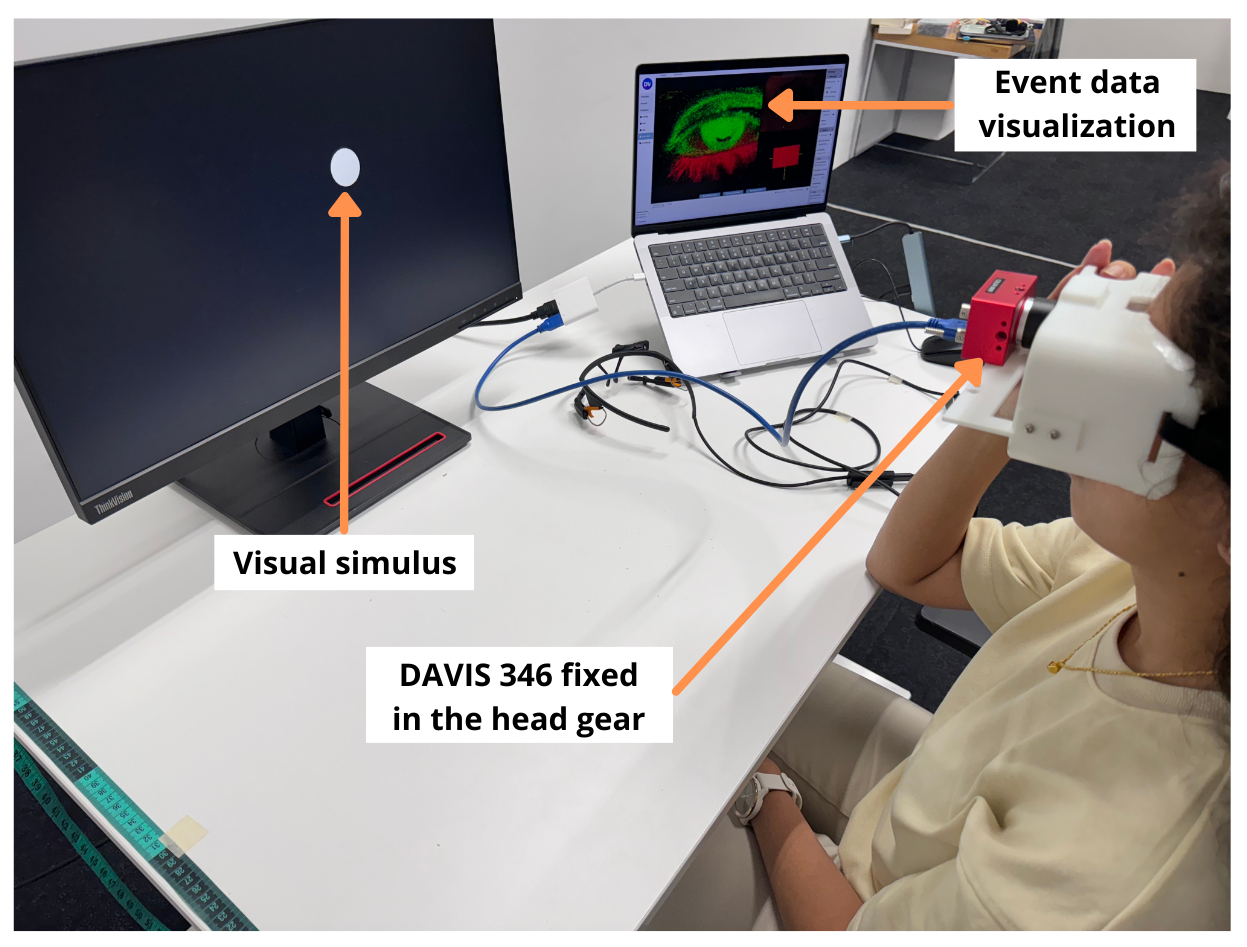}
    \label{fig: setUp_davis}
    }
    \subfigure[]{
    \includegraphics[width=0.45\textwidth]{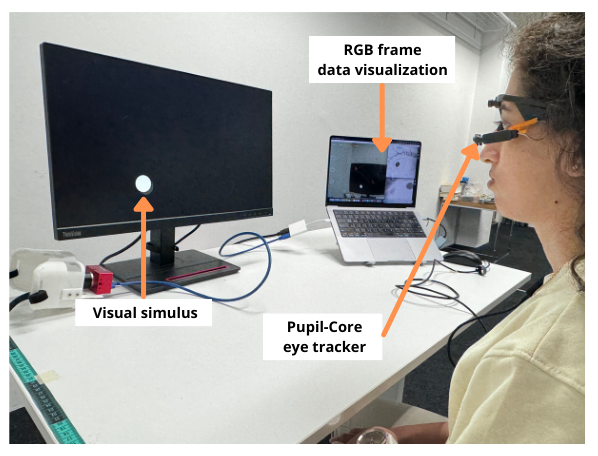}
    \label{fig: setUp_pupil}
    }
    \caption{Data collection setup using (a) DAVIS$346$, (b) Pupil-Core eye tracker.}
    \label{fig:experimentalSetup}
\end{figure}

\section{Datasets, User Studies and Performance Metrics}
\label{sec:dataset}

We now describe our evaluation methodology, which consists of both using (i) pre-existing, ground truth-annotated datasets to evaluate the fidelity of \ourmethod in terms of tracking low-level pupil movement feature and (ii) new, in-lab user studies to quantify \ourmethodf performance in terms of continuous user authentication (\changed{an illustrative application of \ourmethod that we shall detail in Section~\ref{sec:authentication}}).  Table~\ref{tab:dataset} summarizes the key differences between the two datasets, Ev-Eye, and \ourmethodnospace, that we shall now describe. \changed{We emphasize that the pre-existing data is collected under fairly artificial conditions with no head movement, whereas our \ourmethod data is collected under more natural, less-restrictive conditions and thus enables us to study a richer set of user artifacts.}

\subsection{Ev-Eye Dataset for Eye Kinematics}
\label{subsec:ev-eye_dataset}
We first use the published Ev-Eye dataset~\cite{zhao2024ev} to evaluate the performance of \ourmethod in terms of its ability to support fine-grained eye movement tracking. Ev-Eye consists of $48$ participants' data collected using the DAVIS$346$ camera that provides both event and RGB (30 FPS) data. For data collection purposes, they used a visual stimulus with a solid red circle (that disappears and reappears at a different randomly chosen spatial location of the screen for 1.5s) displayed on the monitor to guide the gaze movement of the subject. Along with the DAVIS RGB-frames and events, the dataset also includes (i) the reference Point of Gaze (PoG) captured at relatively high frequency ($\sim$100 Hz) using Tobii Glasses, (ii) pupil segmentation/localization on events data, and (c) dense gaze references for eye movements, such as fixation, saccades, and smooth pursuits.

The Ev-Eye dataset is especially useful in evaluating the fidelity of \ourmethodf Pupil Segmentation and Tracking component, as it provides ground truth annotations for the user gaze. In addition, its inclusion of Tobii tracker data will allow us to compare \ourmethodf event camera-based user authentication accuracy against that obtained using high frequency, ground truth eye tracking data. However, the Ev-Eye dataset was collected under highly controlled conditions where each user had their head position fixed and resting on a chin-rest. 



\subsection{\ourmethod User Study \& Dataset}
While evaluation on Ev-Eye helps establish the benefits of novel \ourmethod components, such as dynamic event slicing, we also need to evaluate \ourmethodf efficacy in terms of low-level eyeball tracking and our eventual exemplar application---user authentication--especially under varying task contexts and more diverse environmental conditions (e.g., low lighting levels, more natural macro-head movements). To achieve this objective, we executed a separate data collection user study, comprising 40 participants, in a laboratory setting.  For this study, an individual participant wore only one device (either our event camera-embedded custom headset or the commercial Pupil-Core eye tracker~\cite{kassner2014pupil}) at any one instant, implying that it was not possible to obtain concurrent ground truth for low-level eye kinematics. 

\begin{table}[tbh]
    \centering
    \caption{Summary of the Datasets}\label{tab:dataset}
    \begin{tabular}{|F{0.08\textwidth}|T{0.1\textwidth}|T{0.12\textwidth}|T{0.2\textwidth}|T{0.1\textwidth}|T{0.23\textwidth}|}
    \hline
    \textbf{Dataset} & \textbf{Devices} & \textbf{Data Format} & \textbf{Experiment Setup} & \textbf{End Goal} & \textbf{Ground Truth Annot.} \\ \hline \hline 
    \textbf{Ev-Eye} & 
    DAVIS$346$ & 
    \begin{itemize}[leftmargin=*]
    \vspace{-0.5cm}
    \item Events
    \item Grayscale frames (25 FPS)
    \end{itemize} &
    \begin{itemize}[leftmargin=*]
    \vspace{-0.5cm}
    \item \textbf{Two} Event cameras
    \item Stationary; Near-eye
    \item Fixed distance b/w visual stimuli and the user
    \end{itemize} &
    Gaze tracking &
    \begin{itemize}[leftmargin=*]
    \vspace{-0.5cm}
    \item Point of gaze from Tobii Pro Glasses 3
    \item Pupil segmentation on events data
    \item Annotations for saccades, fixations, and smooth pursuit using gaze references 
    \end{itemize}
    \\ \hline
    \textbf{\ourmethodnospace} & 
    DAVIS$346$ & 
    \begin{itemize}[leftmargin=*]
    \vspace{-0.5cm}
    \item Events
    \item Grayscale frames (25 FPS)
    \end{itemize} &
    \begin{itemize}[leftmargin=*]
    \vspace{-0.5cm}
    \item \textbf{Single} Event camera
    \item Head mounted (wearable); Near-eye
    \item Mobile
    \end{itemize} &
    Eye movement tracking &
    \begin{itemize}[leftmargin=*]
    \vspace{-0.5cm}
    \item Grayscale images (120 FPS) from Pupil Core tracker
    \item Point of gaze from Pupil Core tracker 
    \end{itemize}
    \\ \hline

    \end{tabular}
\end{table}

\changed{As illustrated in Figure \ref{fig:experimentalSetup}, our data collection had two stages. In the first stage (as depicted in Figure \ref{fig: setUp_davis}), the participants wore a custom-built headgear fixed with a DAVIS$346$ camera~\footnote{\url{https://inivation.com/wp-content/uploads/2019/08/DAVIS346.pdf}, Accessed: \today} secured around the forehead using a Velcro fastener. The camera was positioned adjacent to the right eye, while the participants were directed to track the visual stimuli using their left eye. In the second stage (as depicted in Figure \ref{fig: setUp_pupil}), the participants wore the off-the-shelf Pupil-Core eye tracker~\cite{kassner2014pupil} that is widely used by 
the academic research community. The eye tracker uses two near-eye cameras, oriented towards the wearer's eyes, and one world camera, facing outwards, to respectively capture the pupil's view and location, as well as the scene that engages the participant's gaze. By following numerous studies in the literature~\cite{land2000eye,diaz2013real,collins2009predicting,mann2019predictive}, we design our study protocol to elicit natural eye movements: the visual stimulus appears at the top left corner of the screen and then moves continuously in random directions such that the stimuli exhibits nearly-perfect and smooth collisions when it hits an edge of the screen. The stimulus remains consistent across all participants. To guide the gaze movement of the participants, we displayed the visual stimulus on a $1920 \times 1080$, $23.8$-inch monitor. The distance between the monitor and the participant varied between $45cm$ and $50cm$, resulting in a field of view between $56^\circ \times 34^\circ$ and $62^\circ \times 37^\circ$.}

Our sample consists of 40 participants, including 28 males and 12 females, representing diverse ethnic backgrounds \changed{(i.e., from 7 different nationalities)}. Their ages range from 21 to 32 years, with a mean of 26.08 years and a standard deviation of 2.99. The participants had perfect ($20/20$), contact lens-based corrected or corrective spectacles-assisted vision \changed{with the percentages being $47.5\%$, $10\%$ and $42.5\%$ in the participant pool respectively}. To ensure that an individual participant was able to exhibit meaningful eye movement, only nearsighted participants were included \changed{(i.e., if the participants wore corrective spectacles for any other ocular condition, they were excluded from the study). Nearsighted participants were instructed to remove their corrective spectacles during the experiment (after ensuring that the participants can follow the visual stimuli in their field of vision without any difficulties) since (1) the presence of the spectacles affects the fit of the event camera mounted head-gear and (2) the glint or reflections caused by the spectacles potentially affect the event camera's ability to accurately capture the pixel intensity changes. All participants were recruited through university-wide announcements and local community outreach, ensuring a diverse sample in terms of gender and ethnicity. Inclusion criteria required participants to have a normal or corrected-to-normal vision, verified by a pre-study screening whereas the individuals with significant ocular health issues, those unable to use corrective lenses effectively, or those with conditions other than near-sightedness were excluded.} 

\noindent \changed{\textbf{Ethical Considerations}: Our study was approved by the Institutional Review Board (IRB) of our institution, ensuring that it met ethical standards for research involving human subjects. Furthermore, the participants received a monetary incentive for their participation as guided by IRB protocols. All participants were provided with detailed information about the study, including its purpose, procedures, and any potential risks. If the participant felt unable to perform the specified activities, or if the specified activities made the participant feel uncomfortable, the participant was excluded from the study. In addition, participants could choose to withdraw from the study at any time (before or during the data collection) if they felt uncomfortable.}

\color{black}

The participants were first seated comfortably on a chair; before the actual data collection, the wearable devices were calibrated mechanically to ensure an optimal capture of each participant's eye region \changed{through several steps including (1) adjusting the attached Velcro fastener to ensure the proper fit of the head gear containing the event camera, (2) mechanically adjusting the focal length of the event camera lens such that it is optimal to capture the eye movement dynamics with minimal blur, (3) adjusting the positions and angles of three embedded cameras on the sliding arms in Pupil-Core eye tracker to ensure that the expected views are optimally collected and (4) calibrating and validating the Pupil-Core eye tracker using a 5-point calibration paradigm to ensure that the average calibration error is below $1.5^\circ$ as measured by the accompanying Pupil-Labs software~\cite{kassner2014pupil}}. The participants had the freedom to move their heads and bodies as they wished, without the need to maintain fixed positions. Throughout both stages, the visual stimulus consisted of a solid white circle against a black background displayed on the monitor, with a diameter of 80 pixels \changed{(A video recording of the utilized visual stimuli on the screen is presented in the corresponding repository)}.

Each session per participant consists of four trials, each lasting four minutes. In the first two trials, the participants wore the DAVIS$346$ camera; in the last two trials, they wore the Pupil-Core eye tracker. \changed{Between every two consecutive trials, there was a resting period of at least 30 seconds to reduce visual fatigue to the participant.} The randomized movement pattern of the white circle was identical across a cross-device trial pair (spanning both the DAVIS$346$ and Pupil-Core device) but varied between the two trials corresponding to the same wearable device. Due to their endeavor to focus their gaze on the white circle, each participant predominantly exhibited smooth pursuit and fixation states when the circle was moving smoothly, while saccadic states were triggered by the occasional discontinuous ``jump" in the location of the white circle.

\subsection{Key Performance Metrics}
\label{subsec:perform-metric}
We evaluate \ourmethod vs. alternative competitive baselines using multiple metrics that together capture both our primary goal of accurate pupil detection and our secondary, application-level goal of accurate per-user authentication.

For the pupil detection task, we have adopted two key evaluation metrics:
\begin{itemize}[leftmargin=*]
\item \emph{Intersection over Union (IoU)}: This metric, widely employed in pupil region segmentation~\cite{zhao2024ev}, quantifies the overlap between the estimated and ground truth pupil regions.
\item \emph{Dice Coefficient}: This metric, commonly used in eye segmentation tasks~\cite{zhao2024ev}, gauges the similarity between the estimated and ground truth pupil regions. 
\end{itemize}

For eye movement feature-based user authentication \changed{(the illustrative application (detailed in Section~\ref{sec:authentication}) we choose to demonstrate the efficacy of our proposed \ourmethod technique)}, assess performance primarily through the per-user authentication \emph{Accuracy} metric, which counts the number of correct user authentications over all the authentication instances. \changed{We also compute the False Acceptance Rate (FAR) and the False Rejection Rate (FRR), as well as the \emph{Equal Error Rate} (EER), which is a commonly used metric for biometric authentication, representing the point on a Receiver Operating Characteristic (ROC) curve where FAR equals FRR. 
The FAR quantifies the probability of an unauthorized person being incorrectly identified as an authorized user; a low FAR is essential to reduce the risk of unauthorized access. The FRR, also referred to as False Non-Match Rate (FNMR), represents the likelihood of an authorized user being incorrectly rejected by the system. A low FRR is critical for ensuring a positive user experience, particularly in scenarios where the authentication process is frequently used, such as unlocking a mobile device. High FRR can lead to user frustration and reduced system usability.}

\section{\ourmethodf Performance Overview: Pupil Segmentation}
\label{sec:evaluation}
In this section, we first explain the baseline methods we choose to evaluate the efficacy of \ourmethod. Then we provide various quantitative and qualitative analyses to show the superior performance of \ourmethod, specifically on accurate pupil segmentation under various system parameters. Our results are generated on the publicly available Ev-Eye dataset~\cite{zhao2024ev}.


\subsection{Baselines for Pupil Segmentation}
\label{subsec:baseline}
For the pupil segmentation task, we consider the following three 
baselines: 
\begin{enumerate}[leftmargin=*]
    \item \textbf{Ev-Eye~\cite{zhao2024ev}:} A hybrid event and frame-based pupil segmentation method, where, the grayscale images from the event cameras are used to segment the pupil region (using conventional CV methods for segmentation and denoising) to generate the pupil boundaries that can be used as pupil template. 
    In our analysis, we use the pupil template centers (this specific annotation is provided in the Ev-Eye dataset) as one of the baselines.
    \item \textbf{E-Gaze~\cite{li2024gaze}:} This approach on purely event sensing-based pupil segmentation uses a non-parametric statistical method called two-dimensional kernel density estimation (KDE) to locate the centre of two concentric circles, one representing the iris and the other one representing the pupil region, within an event frame.
    
    \item \textbf{DAVIS346-RGB-30Hz (RGB Frame-based pupil extraction):} For fair comparison, we also introduce a purely RGB frame based pupil segmentation baseline, where we re-implemented \ourmethodf pupil tracking techniques (described in \S\ref{subsec:pupil_tracking}) on the RGB frames in the Ev-Eye dataset (more precisely, the grayscale representation of RGB images released in Ev-Eye), captured by a DAVIS$346$ camera. As an initial step, we employ an additional color filtering technique on these grayscale images to isolate regions of interest containing black contours, which typically represent pupils. Following the extraction of black contours, we execute \ourmethodf standard steps of (i) using the Hough technique~\cite{hough} to identify elliptical shapes within these regions of interest, thereby segmenting the boundaries of candidate pupil objects, and then (ii) applying RoI thresholding (on both the area and aspect ratio of the detected contours) to precisely isolate the pupillary contour.

\end{enumerate}

\subsection{Performance of Pupil Segmentation}




\begin{table}[]
    \centering
    \caption{\changed{Performance evaluation of different pupil segmentation methods (Ev-Eye Dataset).}}\label{tab:eval}
    \begin{tabular}{|c|c|c|c|c|}
    \hline
    \textbf{Methods}                & \textbf{MAE} & \textbf{IoU (\%)} & \textbf{Dice Coeff. (\%)} & \textbf{Latency (s)} \\ \hline
    \textbf{\ourmethod}                & 10.13        & 92                & 89                        & 0.0047               \\ \hline
    \textbf{Ev-Eye}                 & 14           & 89                & 88                        & 0.71                 \\ \hline
    \textbf{DAVIS346-RGB-30Hz}         & 16           & 84                & 81                        & 0.008                 \\ \hline
    \textbf{E-Gaze (Fixed 2000 \# of events)}                 & 26           & 48                & 44                        & 0.012                \\ \hline
    \textbf{E-Gaze with adaptive slicing}                 & 12           & 87                & 85                        & 0.012                \\ \hline
    \end{tabular}
\end{table}

\begin{figure}
    \centering
    \subfigure[]{
    \includegraphics[width=0.45\textwidth]{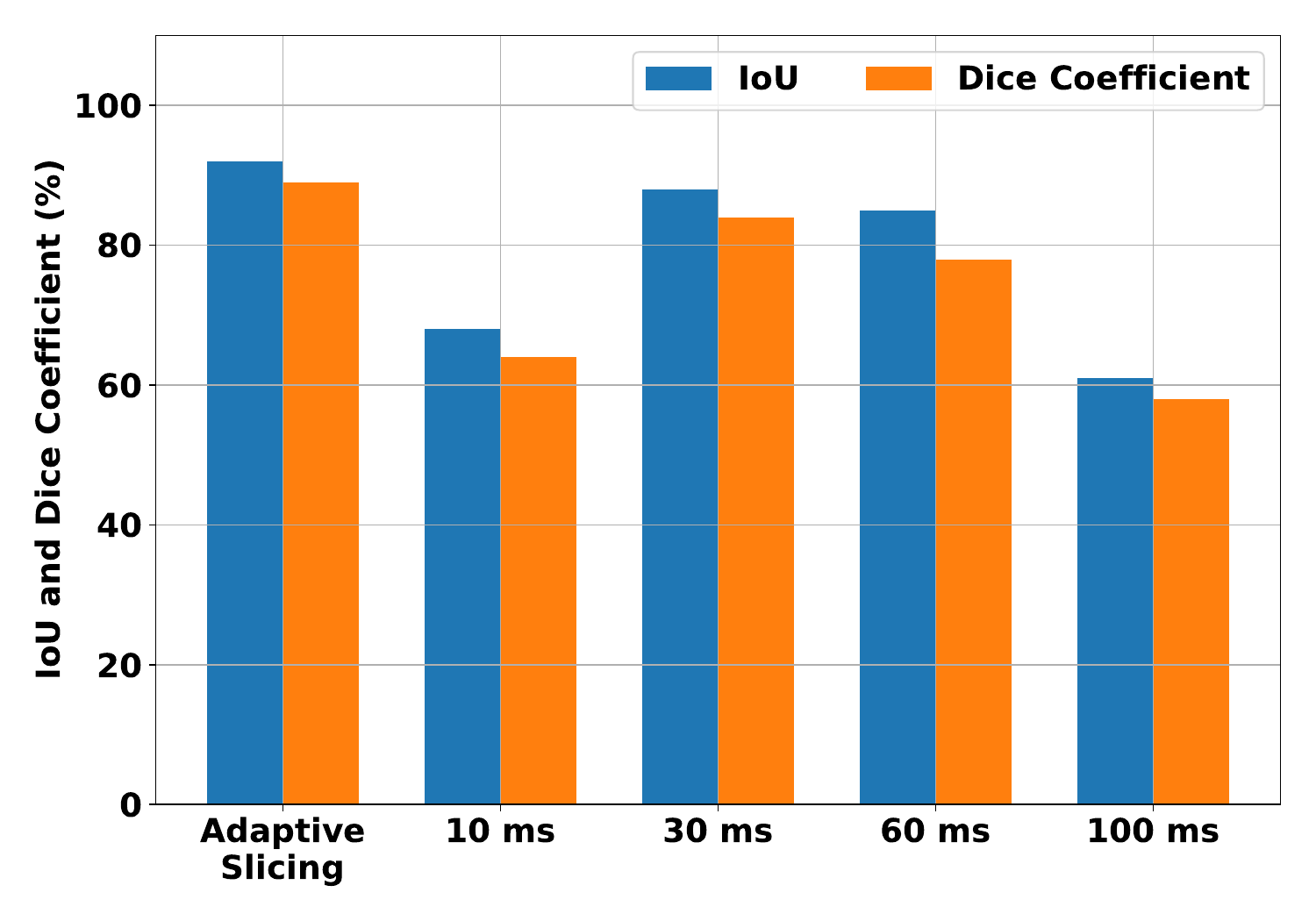}\label{fig:slicing_pupil}
    }
    \subfigure[]{
    \includegraphics[width=0.45\textwidth]{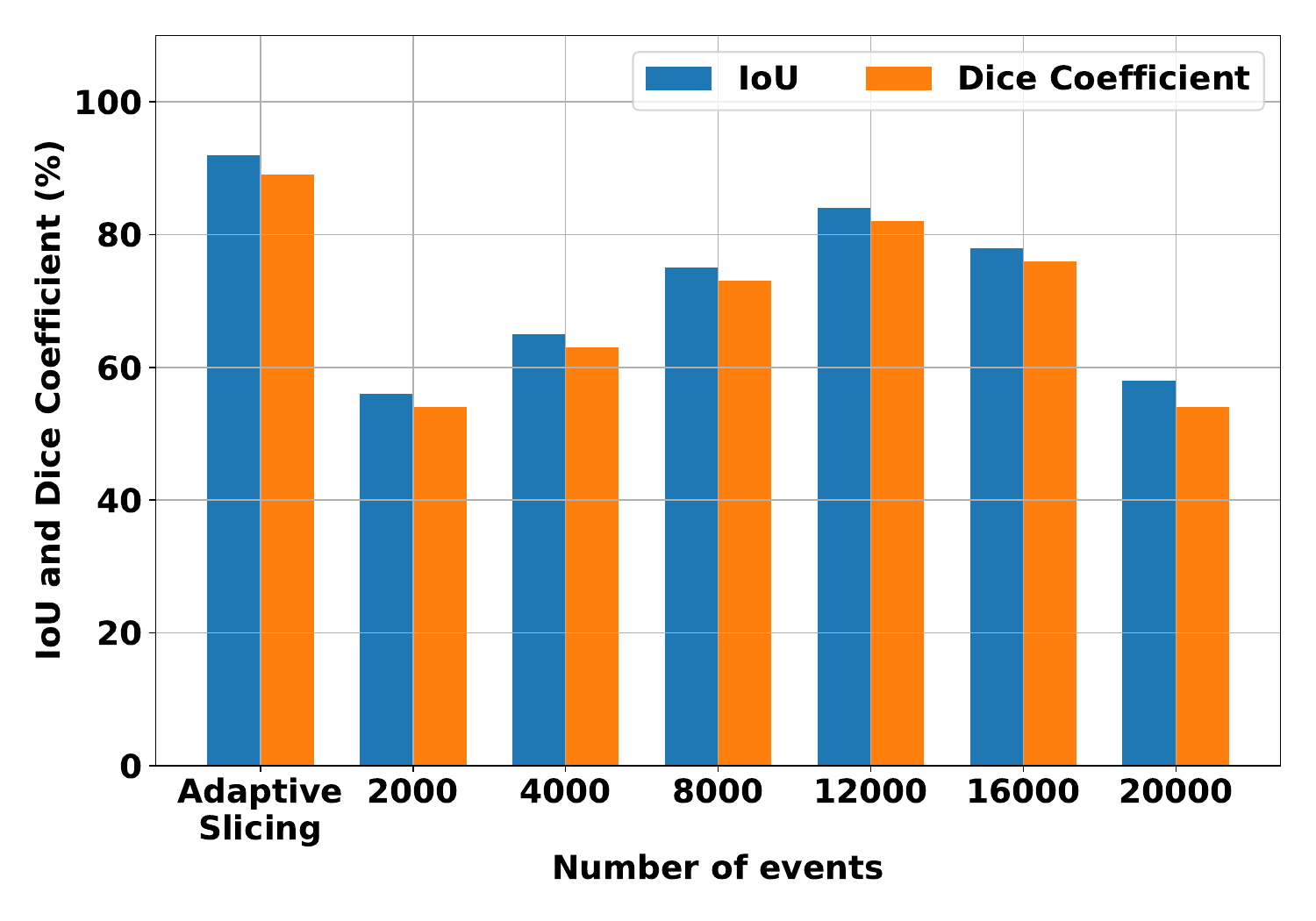}\label{fig:num_events}
    }
    \caption{IoU and Dice Coefficients comparison of \ourmethod against various (a) fixed-time-based framed representations, (b) fixed-event volume-based framed representations.}
\end{figure}

As summarized in Table~\ref{tab:eval}, \ourmethod{} achieved higher IoU ($\approx 92\%$) and dice coefficient ($\approx 89\%$) compared to other methods on the Ev-Eye dataset. Please note that \ourmethodf adaptive slicing achieves the nominal frame rate of 30 FPS. \changed{The Ev-Eye and E-Gaze (with our proposed adaptive slicing for event accumulation) also demonstrated competitive performance with an IoU of $89\%$ and $87\%$, respectively, but as we shall see shortly, their computation load is much higher. Note, E-Gaze is also a purely event-based approach and it considers a fixed event volume of 2000 events to form the framed representation~\cite{li2024gaze}, however, we have demonstrated how varying the slicing approach can impact the pupil segmentation performance later in Section~\ref{sec:egaze_per}.} The purely frame-based approach (evaluated using the DAVIS RGB-frames from the Ev-Eye dataset) has a lower accuracy as the pupil segments are generated using a standard image processing-based approach, which can sometimes lead to false positives in pupil detection. Also, as the Ev-Eye frame data is captured at a lower temporal resolution ($\approx 30$ FPS) with the DAVIS$346$ camera, the captured frame can have unstable motion artifacts during periods of rapid eye motion, leading to poorer accuracy compared to the event-based approaches where the intensity of the data stream increases in proportion to the velocity of eye movement.

\subsubsection{Pupil segmentation under different slicing windows}
To demonstrate the benefits of adaptive slicing, we next evaluate the performance of \ourmethod vs. alternatives that utilize a fixed window (of either time or event count). Our results, shown in \figurename~\ref{fig:slicing_pupil}, indicate that adaptive slicing outperforms fixed slicing windows across different scenarios. For instance, when using a $10$ ms slicing window, the IoU is $68\%$ and the Dice Coefficient is $64\%$. However, with adaptive slicing, the IoU increases to $92\%$ and the Dice Coefficient to $89\%$. For low values of the slicing window and slow eye movement (fewer events), the eye region may not be fully captured in the framed representation, leading to lower accuracy. Conversely, using a larger slicing window with more events may result in multiple overlapping pupil regions captured in a single event frame, introducing more noise and reducing the accuracy of the pupil region detection.


\subsubsection{Pupil Segmentation under Different Number of Events}\label{sec:num_events}
We also analyze the impact of varying numbers of events on pupil segmentation accuracy. The number of events is varied from $2000$ to $20,000$, and the segmentation accuracy is compared with an adaptive slicing-based approach. Figure~\ref{fig:num_events} illustrates that the IoU and Dice Coefficient values demonstrate superior performance when the number of events is greater than $8000$ and less than $16000$. However, in scenarios with lower eye movement between two rapid eye movements, using a fixed number of events can result in an unstable framed representation. This instability is characterized by the accumulation of two random pupil regions generated at the two extreme timestamps of the slice cut, with intermediate minor events generated due to lower eye movements. In contrast, adaptive slicing waits until it detects a rapid eye motion before slicing out the events, leading to a more stable framed representation. Consequently, adaptive slicing achieves superior IoU and Dice coefficients compared to fixed-number-of-events-based slicing.

\begin{figure}
    \centering
    \subfigure[]{
    \includegraphics[width=0.45\textwidth]{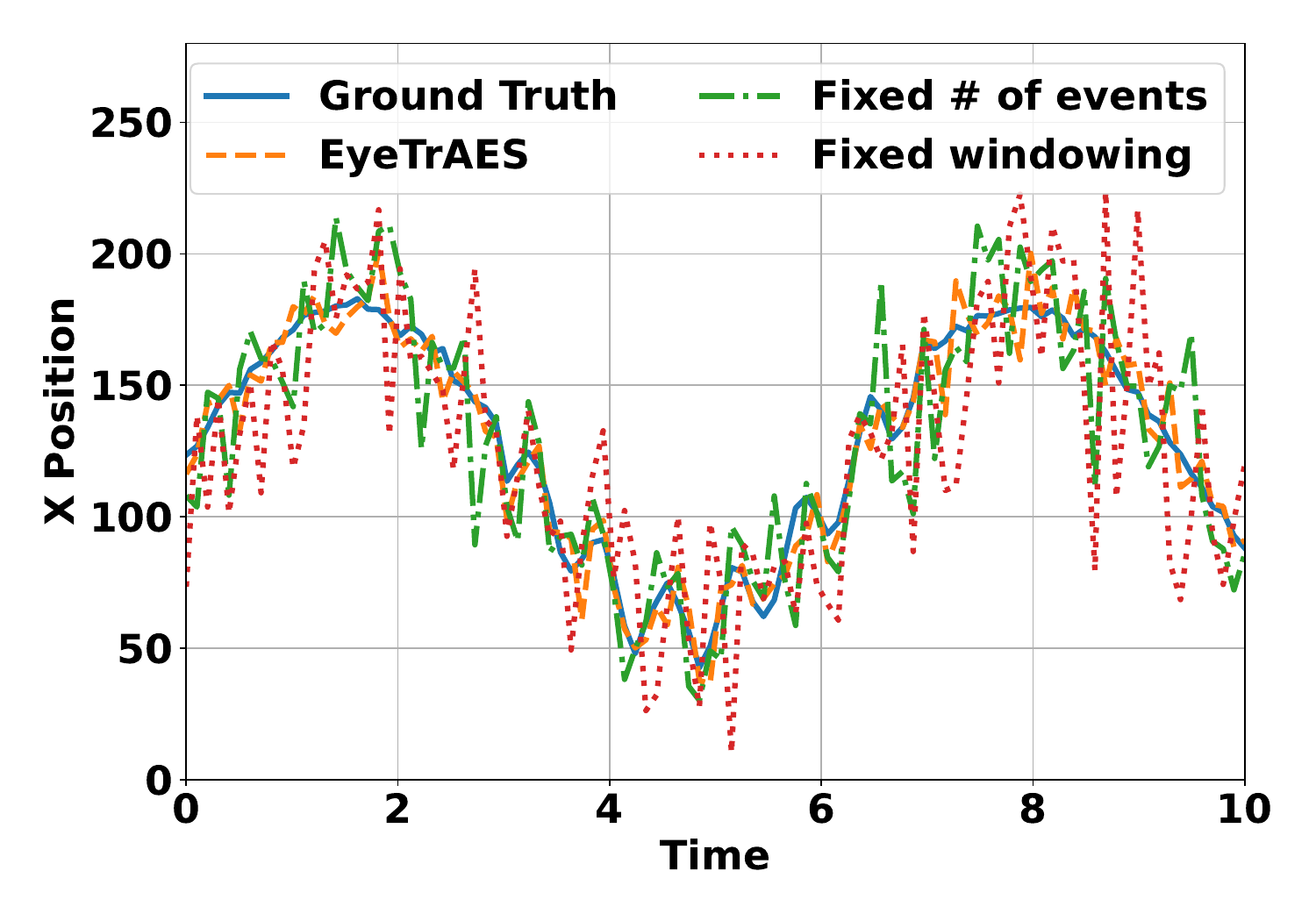}\label{fig:x_pupil}
    }
    \subfigure[]{
    \includegraphics[width=0.45\textwidth]{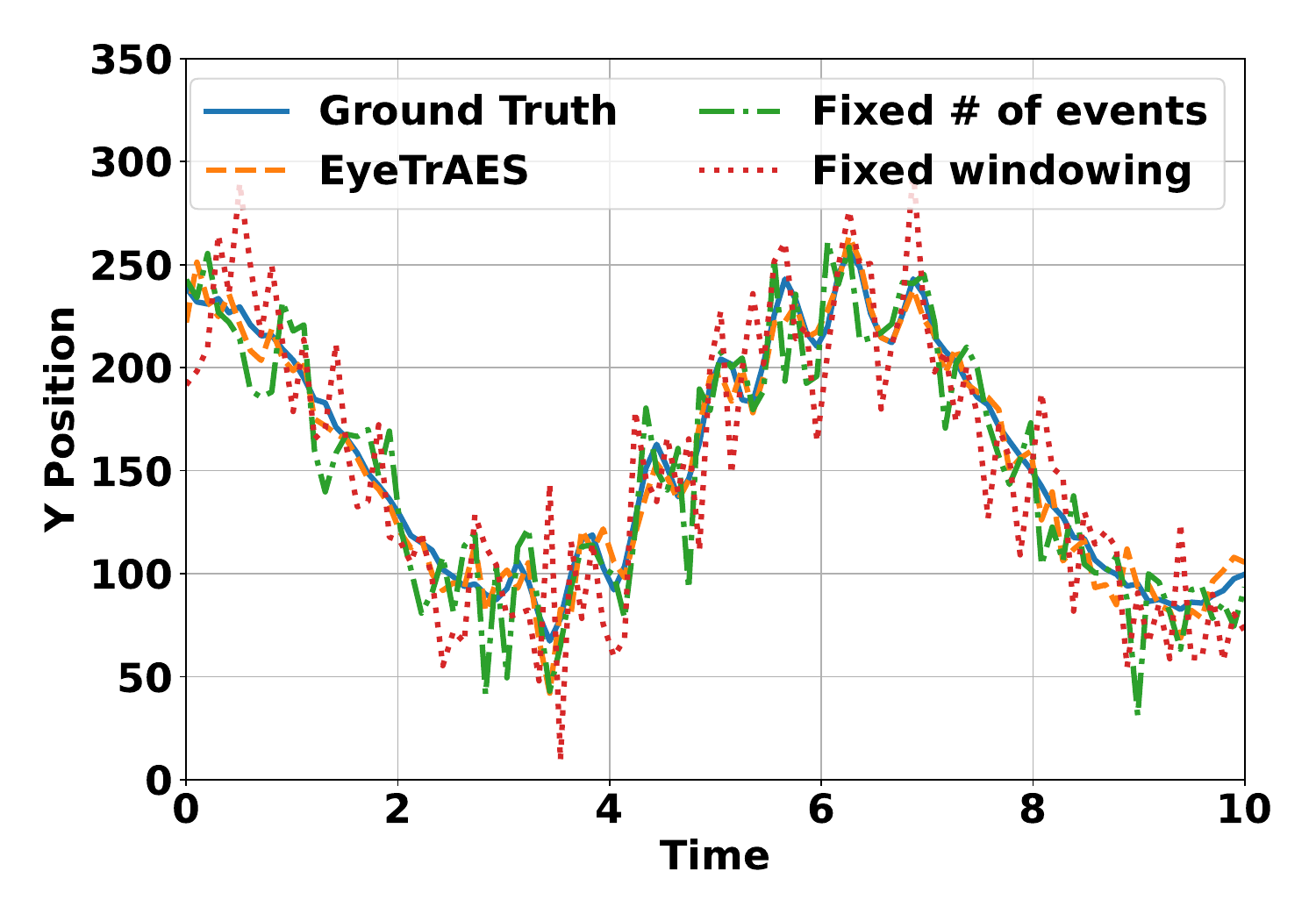}\label{fig:y_pupil}
    }
    \caption{\changed{Comparison of pupil coordinates over time using different slicing strategies. The figure illustrates the performance of adaptive event slicing (AES) versus fixed-number-of-events (8000–16000) and fixed slicing window lengths (30 ms). AES consistently provides a more accurate representation of pupil coordinates, effectively capturing dynamic eye movements.}}
    \label{fig:x_y_pupil}
\end{figure}

\changed{To further substantiate our findings, we conducted a qualitative evaluation using 10 seconds of randomly selected data from the dataset, comparing pupil coordinates over time using our adaptive slicing method against fixed-number-of-events (ranging from 8000 to 16000) and a fixed slicing window length of 30 ms. The results, illustrated in \figurename~\ref{fig:x_y_pupil}, reveal that our adaptive slicing method consistently outperforms both fixed slicing strategies. The adaptive approach more accurately captures dynamic pupil movement and adjusts to varying eye motion rates, thereby providing a more precise representation of pupil coordinates over time. Notably, while fixed-number-of-events approaches demonstrate slightly better performance than fixed slicing windows of 30 ms, the improvements are marginal compared to the significant gains achieved with adaptive slicing. This qualitative analysis, based on a randomly selected 10-second segment of the dataset, underscores the advantages of our adaptive method in maintaining high accuracy and stability across diverse eye movement scenarios.}

\subsubsection{Computation Latency}
As demonstrated  in Table~\ref{tab:eval}, \ourmethod is also  superior in terms of segmentation latency  compared to all the baselines, incurring an average computational latency of 4.7 ms. In contrast, the closest  baseline, DAVIS346-RGB-30Hz, takes almost twice as long (average=8 ms), as it involves multiple image processing steps including color based filtering, morphological operations and contour detection on a denser pixel representation; note that the IoU of DAVIS346-RGB-30Hz is $\sim$10\% lower than that of \ourmethodnospace. For E-Gaze, the latency is almost $3\times$ larger (average$\sim=$12 ms), as it utilizes two separate concentric circle fitting steps (for both the pupil and iris regions), while \ourmethod applies a Hough based circle detector only once. Not surprisingly, Ev-Eye has the largest latency (average$\sim$=710 ms, almost $150\times$ of that of \ourmethodnospace), as it first uses a computationally complex U-Net model  to identify candidate pupil segments, followed by additional candidate point subset estimation to filter out noisy events caused by the movement of eyelashes and eyelids.

\subsubsection{\changed{Impact of Event Slicing on other Pupil Segmentation Baselines}}\label{sec:egaze_per}

\begin{figure}
    \centering
    \subfigure[]{
    \includegraphics[width=0.45\textwidth]{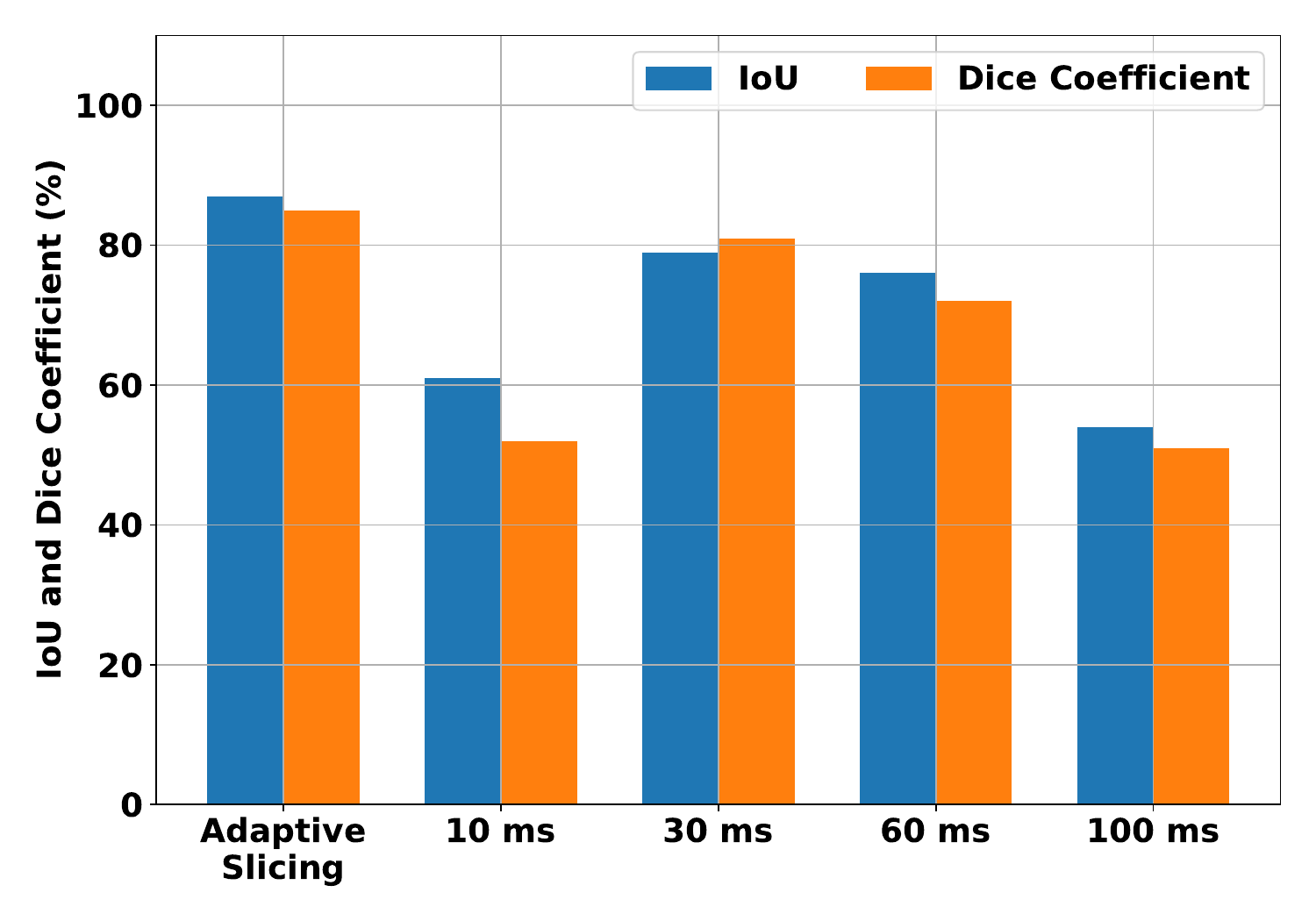}\label{fig:eaze_slicing}
    }
    \subfigure[]{
    \includegraphics[width=0.45\textwidth]{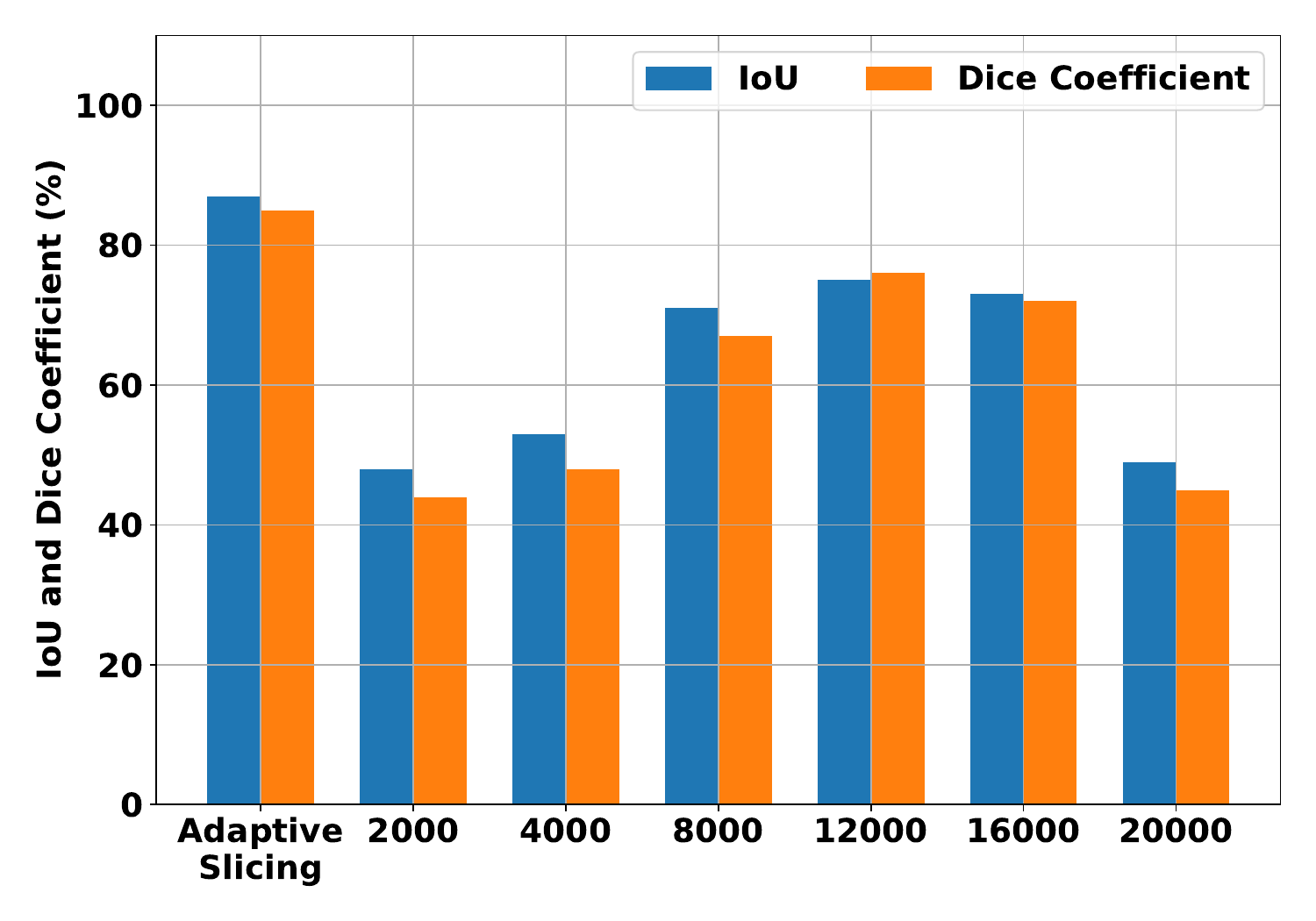}\label{fig:egaze_num}
    }
    \caption{\changed{Performance of E-Gaze with different slicing strategies: (a) with different slicing window length, (b) with different number of events}}\label{fig:egaze_perform}
\end{figure}


\changed{To evaluate the effectiveness of our proposed adaptive slicing technique in conjunction with an existing pure event-based pupil segmentation baseline such as E-Gaze~\cite{li2024gaze}, we conducted a series of experiments. These experiments involved varying the fixed slicing window lengths or the fixed volumes of events. We then compared the IoU and Dice coefficients for pupil segmentation using these methods against our adaptive slicing strategy. As depicted in \figurename~\ref{fig:egaze_perform}, our adaptive slicing method significantly improves event aggregation within the framed representation, resulting in superior IoU and Dice coefficients compared to fixed slicing strategies. Notably, E-Gaze captures 2000 events to form the framed representation~\cite{li2024gaze} which performs the poorest on the Ev-Eye dataset.}


\changed{The results collectively demonstrate not just the superiority of our combined adaptive slicing and lightweight pupil segmentation techniques, but also show that adaptive event slicing helps improve the tracking accuracy of other prior computationally-heavier segmentation baselines. After witnessing how our combined adaptive event slicing and light-weight pupil segmentation technique achieves highly accurate and low-latency pupil localization (segmentation), we shall now explore how \ourmethod allows us to extract high temporal resolution pupillary kinematics to support improved biometric user authentication (as an exemplar application).}
\section{Exemplar Application: \ourmethodnospace-based User Authentication}
\label{sec:authentication}

We now proceed to show how \ourmethod-based pupil segmentation and tracking can be used to provide improved user authentication. Our key hypothesis is that the fine-grained micro-movements of the pupil, that naturally occur during regular viewing activities, vary across individuals (due to variations in ocular muscle strength), and can thus serve as a biometric fingerprint.


\begin{figure*}[thb]
        \centering
        \vspace{-0.1in}
        \includegraphics[width=0.6\textwidth]{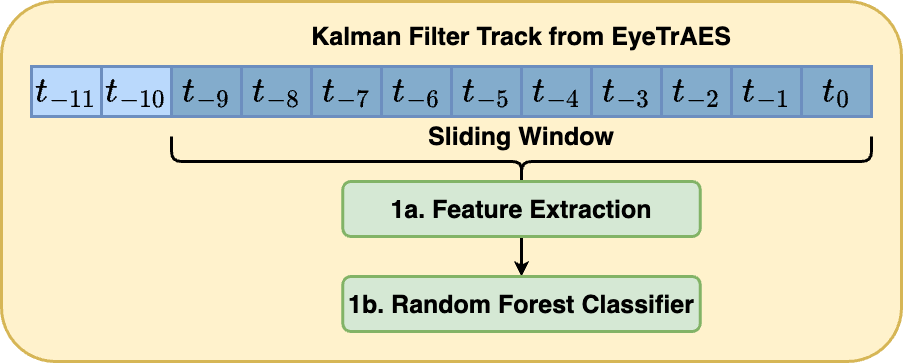}
        \caption{\ourmethodnospace-based User Authentication: Block Diagram of Sub-Components}
        \label{fig:big_picture}
        \vspace{-0.1in}
\end{figure*}

\begin{figure}
    \centering
    \includegraphics[width=0.7\textwidth]{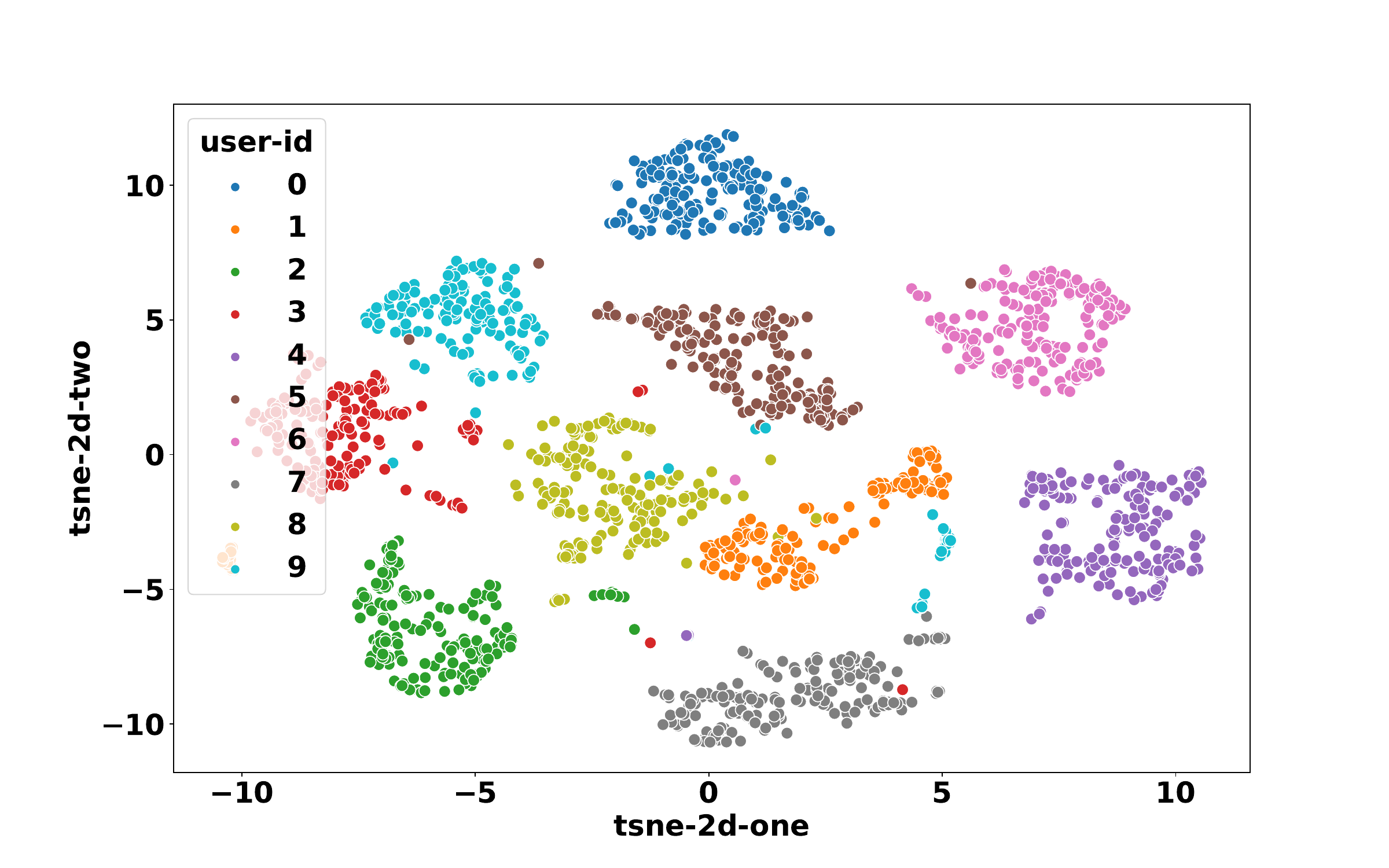}
    \caption{t-SNE distribution of the computed features across 10 subjects from the Ev-Eye dataset in 2D space.}
    \label{fig:tsne}
\end{figure}

\subsection{Pupil Movement Feature Extraction for User Authentication}
As discussed in previous works~\cite{zhang2012biometric,al2022extensive}, individuals often exhibit distinctive saccadic and micro-saccadic eye movements which can be used to authenticate users. However, to compute saccades or micro-saccades, one must have the actual gaze location or the screen coordinate on the screen. Our approach of near-field single eye tracking cannot provide such a gaze estimate. Instead, in the final component of \ourmethodf pipeline, illustrated in Figure~\ref{fig:big_picture}
, we utilize the segmented, Kalman-filtered sequence of pupil locations in a sliding window of size=10 to extract two additional features related to the kinematics of \emph{pupil movement}, using such features as a microscopic proxy for gaze-related artifacts such as saccades, described next. 

\begin{enumerate}[leftmargin=*]
\item \emph{Pupil Velocity}: From the tracked pupil coordinates (say $x_i, y_i$ at time $t_i$), we first compute the first-order derivative of the pupil coordinates $(v_{x,i}, v_{y,i}) = (\frac{x_{i} - x_{i-1}}{t_{i} - t_{i-1}}, \frac{y_{i} - y_{i-1}}{t_{i} - t_{i-1}})$, signifying the velocity or the relative change in the pupil coordinates. This derivative provides an approximation to saccadic or fixating movements of the eye. For example, while a user has a saccadic eye movement, the relative change in the successive pupil coordinate will be much higher than fixation. 
\item \emph{Pupil Acceleration}: We then compute the second order derivative of the pupil coordinates, deriving acceleration values $(a_{x,i}, a_{y,i}) = (\frac{v_{x,i} - v_{x,i-1}}{t_{i} - t_{i-1}}, \frac{v_{y,i} - v_{y,i-1}}{t_{i} - t_{i-1}})$.
\end{enumerate}

\subsection{Feature Vector and Random Forest Classifier} 
The process above creates a tuple of pupil (velocity, acceleration) values for each pair of consecutive event frames--i.e., for the $i^{th}$ frame, we obtain not only the pupil position $(x_i, y_i)$, but also the velocity $(v_{x,i}, v_{y,i})$ and acceleration $(a_{x,i}, a_{y,i})$ values. To clasify an individual user, we then concatenate $M$ consecutive (position, velocity, acceleration) triples, creating an $M\times3$-dimensional feature vector representing \emph{microscopic pupillary motion} attributes over relatively short time windows. Our \ourmethod implementation uses an empirically derived value of $M=$10, effectively representing pupil movement-related features predominantly over 100-400 ms time windows.  This feature vector is then input to a Random Forest classifier with $100$ decision trees, which is trained in a supervised fashion to support binary (one-vs.-rest), per-person classification. 

While we defer detailed evaluation of authentication accuracy till later, we now present initial results that validate our hypothesis about the distinctiveness of pupillary kinematic features, such as velocity and acceleration. We compute these features across 10 selected subjects from the Ev-Eye dataset~\cite{zhao2024ev} and study the t-distributed stochastic neighbour embedding (t-SNE)~\cite{van2008visualizing}, a statistical method for visualizing high-dimensional feature distribution in a lower-dimensional space (with 2 dimensions in this case). The `X' and `Y' axes indicate the first and second dimensions resulting from the dimensionality reduction process. As observed from the~\figurename~\ref{fig:tsne}, different subjects have distinctive, \emph{non-overlapping} distributions of the features, \emph{strongly suggesting that these features can be used to authenticate individual subjects}. 



\section{\ourmethodf Performance: User Authentication}
\label{sec:implementation}
In this section, we evaluate \ourmethodf performance on user authentication on both Ev-Eye and \ourmethod datasets. While we report the overall aggregate performance results using Ev-Eye dataset, we use \ourmethod dataset to (a) performed more detailed studies on the sensitivity to various framing/slicing techniques, and (b) study the impact of various contextual/ambient conditions on the authentication accuracy.

\subsection{User Authentication Performance on Ev-Eye Dataset}
\begin{figure}
    \centering
    \subfigure[]{
    \includegraphics[width=0.45\textwidth]{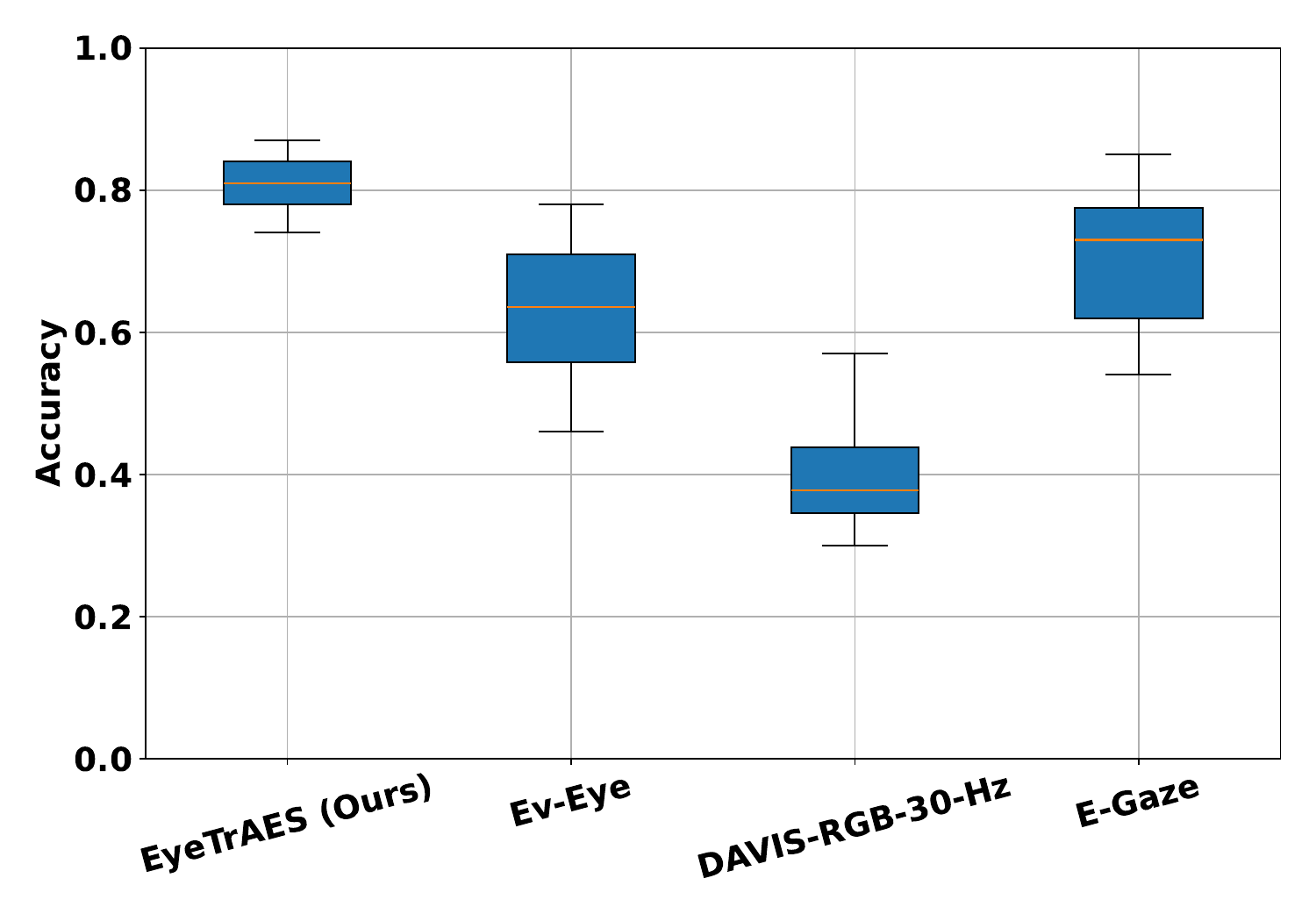}\label{fig:acc_diff_method}
    }
    \subfigure[]{
    \includegraphics[width=0.45\textwidth]{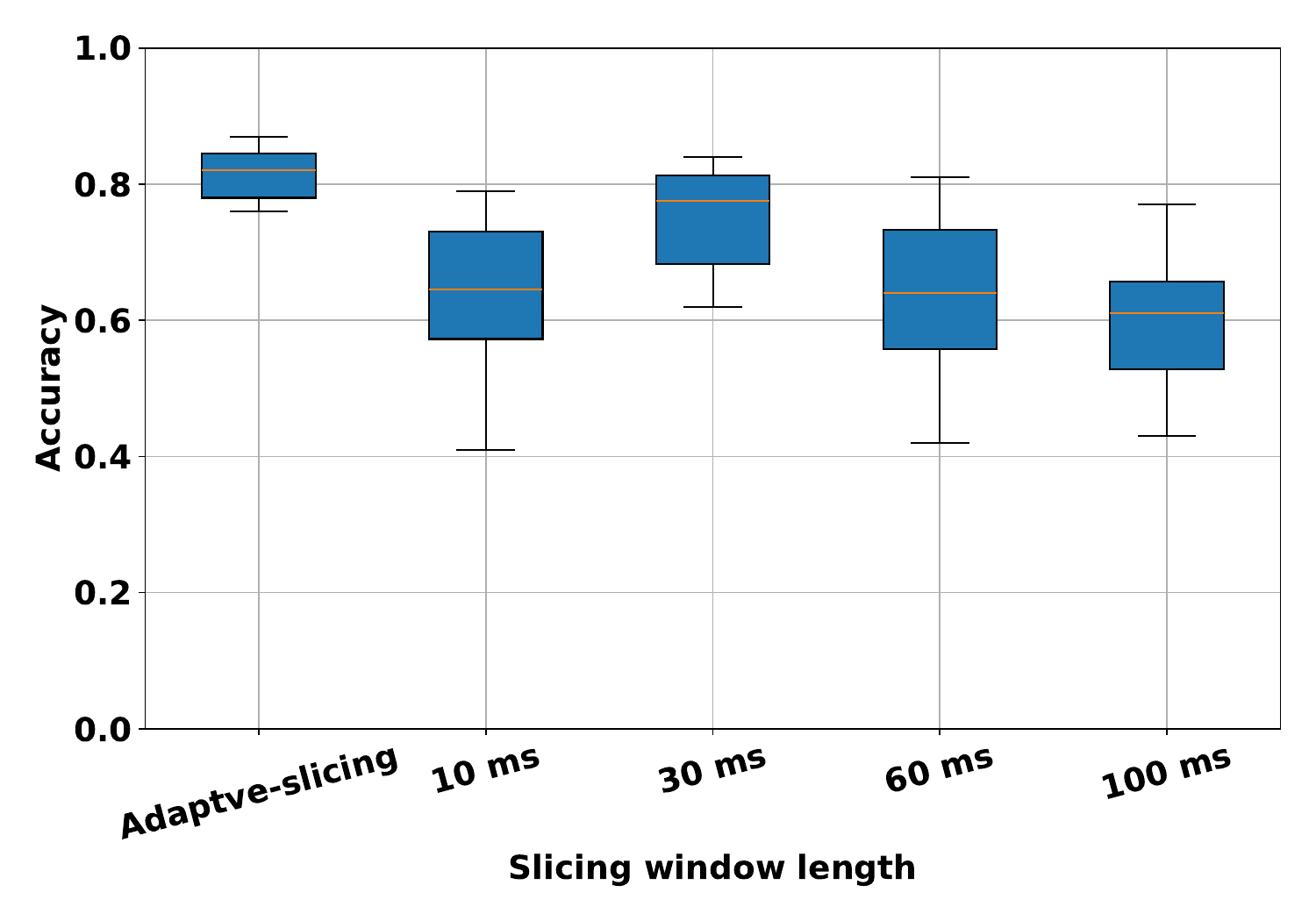}\label{fig:acc_diff_slices}
    }
     \subfigure[]{
    \includegraphics[width=0.45\textwidth]{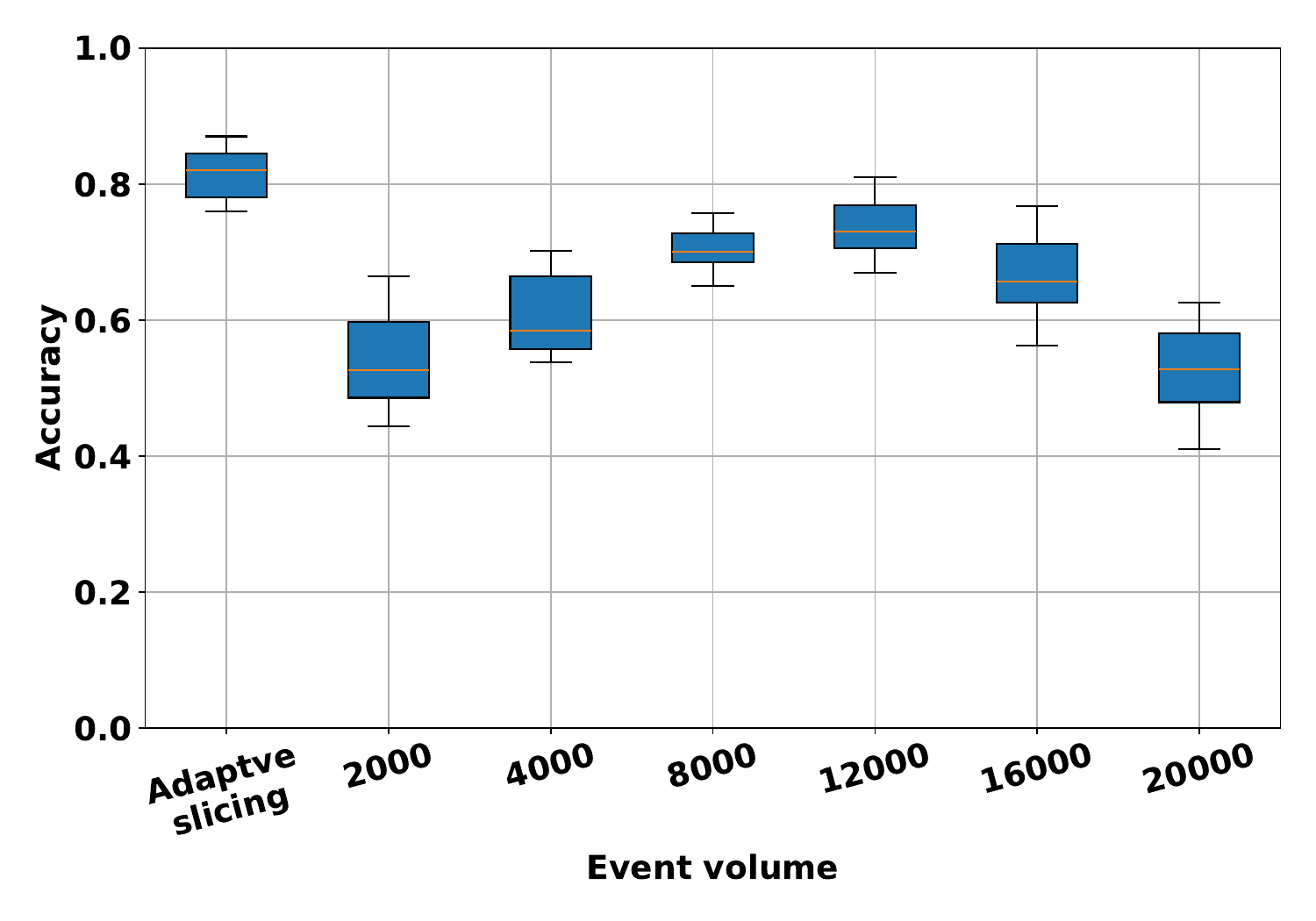}\label{fig:acc_diff_event}
    }
    \caption{User authentication accuracy evaluated on Ev-Eye dataset (a) under different approaches, (b) for the event-based approach with different slicing windows, (c) for different numbers of event accumulation.}
\end{figure}

We conduct a comprehensive comparison of the accuracy achieved by different baselines for eye movement feature-based user authentication, as depicted in \figurename~\ref{fig:acc_diff_method}. \ourmethodf performance is superior to all baselines, achieving an accuracy between $0.78$ to $0.87$, with an impressive median accuracy of $0.82$. Both Ev-Eye and frame-based methods rely on the grayscale frame data for pupil segmentation captured at a lower temporal granularity at 30 FPS. They are thus unable to capture the fine-grained saccadic and micro-saccadic eye movement features which we believe to be key components of the biometric fingerprint, and consequently have lower accuracy compared to \ourmethod{} and E-Gaze, both of which utilize event frames.
While \ourmethod relies on adaptive slicing-based framed representation generation, E-Gaze relies on a fixed-event volume based framing. Thus, both these methods are able to broadly generate the eye movement features in sync with the dynamics of the eye movement, and consequently provide superior accuracy over frame-based pupil segmentation approaches. However, for E-Gaze, the accuracy in segmenting out the pupil region is poor as it identifies the pupil segment only when it captures \emph{both} the concentric circles of the iris and pupil. These findings highlight the effectiveness of our proposed method in achieving superior performance in eye movement-based user authentication scenarios. 

\subsubsection{Performance of event-based approach under different fixed-time slicing window}
For the event-based approach, the slicing technique we use has a significant impact on the ability to estimate the kinematics of the pupil movement, thereby affecting the overall accuracy. This effect is demonstrated in Figure \ref{fig:acc_diff_slices}, where the impact of adaptive slicing on authentication accuracy is compared against different motion agnostic fixed-time windows.

In Figure \ref{fig:acc_diff_slices}, we see that adaptive slicing method achieves significantly higher accuracy, followed by slicing periodically at every 30ms intervals. Slices of shorter (i.e, 10ms) and longer (e.g., 100ms) achieve lower accuracy: a smaller slicing window of 10ms can result in very sparsely populated event frames under slower eye movement, whereas a larger slicing window of 30ms or higher can cause fast eye movements to overlap, leading to noise in the computed saccadic and micro-saccadic eye features. 


\subsubsection{Performance of event-based approach under different fixed-event volume slicing}
We now evaluate the impact of adaptive slicing on authentication accuracy and compare the performance against different fixed-event volume slices. As shown in Figure \ref{fig:acc_diff_event}, the overall accuracy of fixed-event volume slicing is lower compared to adaptive slicing based framed representation. The primary reason behind this can be attributed to the fact that, similar to the fixed-time window, the fixed event volume-based pupil segmentation approach also fails to accurately capture the pupil kinematics, as discussed in Section~\ref{sec:num_events}. The highest median accuracy is achieved when an individual event frame accumulates  12000 events; not surprisingly, this value is where the pupil segmentation accuracy is also much more accurate (Figure~\ref{fig:acc_diff_event}).

\subsubsection{Comparison of Classifier Performance for User Authentication:}
\changed{In this section, we compare the performance of various classifiers for user authentication, focusing on both accuracy and latency. We evaluated the following classifiers: Random Forest, Support Vector Machines (SVM), Radial Basis Function (RBF) Networks, and Gradient Boosting Trees. Our goal was to identify the classifier that provides the best balance of accuracy and latency for real-time applications.}

\changed{The classifiers were trained and tested on the Ev-Eye dataset comprising position, velocity, and acceleration features of the selected pupil region. The performance metrics used for comparison were accuracy and latency.}

\begin{table}[]
\centering
\caption{\changed{Accuracy and latency of user Authentication across different classifiers.}}
\label{tab:classifier}
\begin{tabular}{|c|c|c|}
\hline
\textbf{Classifier}        & \textbf{Median Accuracy (\%)} & \textbf{Latency (ms)} \\ \hline
\textbf{Random Forest}     & 82            & 12           \\ \hline
\textbf{SVM}               & 76.4          & 20           \\ \hline
\textbf{RBF Network}       & 78.1          & 25           \\ \hline
\textbf{Gradient Boosting} & 79.3          & 15           \\ \hline
\end{tabular}
\end{table}

\changed{From the results as shown in Table~\ref{tab:classifier}, we can observe that the Random Forest classifier outperforms the other approaches in terms of both accuracy and latency. The higher accuracy and lower latency make Random Forest the most suitable choice for real-time user authentication applications. The Random Forest classifier achieved an median accuracy of $82.0\%$ and the lowest latency of 12 ms. Its ability to handle high-dimensional data and provide robust predictions makes it an ideal choice for this application. The SVM classifier achieved an accuracy of $82.5\%$ with a higher latency of 20 ms. While SVMs are effective for classification tasks, their higher computational cost makes them less suitable for real-time applications compared to Random Forest. The RBF Network showed an accuracy of $80.3\%$ and a latency of 25 ms. Despite its ability to model complex relationships, its performance was not competitive with the other classifiers. The Gradient Boosting classifier achieved an accuracy of $83.1\%$ with a latency of 15 ms. While its performance was close to that of Random Forest, the slightly higher latency made it less favorable for real-time applications.}

\subsubsection{Feature Ablation Study}
\changed{To understand the contribution of different features to the performance of our user authentication system, we conducted a feature ablation study. We evaluated the model's accuracy by considering various combinations of position, velocity, and acceleration vectors of the eye pupil region. This study helps to highlight the importance of each feature and their combined effect on the model's performance. We tested the following exhaustive feature combinations: (i) Position vectors only; (ii) Velocity vectors only; (iii) Acceleration vectors only; (iv) Position and velocity vectors; (v) Position and acceleration vectors; (vi) Velocity and acceleration vectors; and (vii) Position, velocity, and acceleration vectors.}
\begin{table}[]
\centering
\caption{\changed{Feature abalation study}}\label{tab:abalation}
\begin{tabular}{|c|c|}
\hline
\textbf{Feature Combination}                           & \textbf{Median Accuracy (\%)} \\ \hline
\textbf{Position only}                        & 44.5          \\ \hline
\textbf{Velocity only}                        & 53.4          \\ \hline
\textbf{Acceleration only}                    & 55.7          \\ \hline
\textbf{Position and Velocity}                & 59.2          \\ \hline
\textbf{Position and Acceleration}            & 58.3          \\ \hline
\textbf{Velocity and Acceleration}            & 71.4          \\ \hline
\textbf{Position, Velocity, and Acceleration} & 82            \\ \hline
\end{tabular}
\end{table}

\changed{As shown in Table~\ref{tab:abalation} using only position vectors yielded a median accuracy of $44.5\%$. While position data provides basic information about eye movements, it lacks the dynamic aspects captured by velocity and acceleration. Considering only velocity vectors improved the accuracy to $53.4\%$. Velocity captures the rate of change in position, providing more insight into the movement dynamics. Using acceleration vectors alone resulted in an accuracy of $55.7\%$. Acceleration captures changes in velocity, adding another layer of dynamic information. }

\changed{Combining position and velocity vectors resulted in an accuracy of $59.2\%$. The addition of velocity data to position vectors significantly improved the model's performance. The combination of position and acceleration vectors yielded an accuracy of $58.3\%$. While better than using position alone, it was slightly less effective than combining position and velocity. Using both velocity and acceleration vectors improved the accuracy to $71.4\%$. This combination captures both the rate of change and the changes in the rate of change, providing a richer representation of pupil movements.}

\changed{The best performance was achieved by combining all three features, with a median accuracy of $82.0\%$. This indicates that higher-order pupil movement features, such as velocity and acceleration, significantly enhance user authentication performance.}

In the following section, we further investigate and provide more in-depth analyses on user authentication accuracy using our own \ourmethod dataset.

\subsection{User Authentication Performance on \ourmethod Dataset}
\label{subsec:authperf}
In this section, we discuss the overall user authentication accuracy evaluated on our \ourmethod dataset. We use the same set of baselines described in \S\ref{subsec:baseline}, with the slight modification for \textbf{Ev-Eye} where we use the published Ev-Eye U-Net-based pupil segmentation model to re-train on \ourmethod dataset for pupil segmentation and localization. In addition to the previous baselines, we include we add a few variations of \textbf{DAVIS346-RGB-30Hz} baseline. In particular, we leverage the grayscale images captured by the Pupil Core eye tracker at the nominal frame rate of 120 FPS, and create additional baselines at the down-sampled frame rate of 30, 60, and 90 FPS. These additional baselines will help us understand the efficacy of RGB-based methods in user authentication at higher frame rate, as opposed to the proposed event-based \ourmethod approach which has a lower nominal frame rate.


\begin{figure}
    \centering
    \subfigure[]{
    \includegraphics[width=0.45\textwidth]{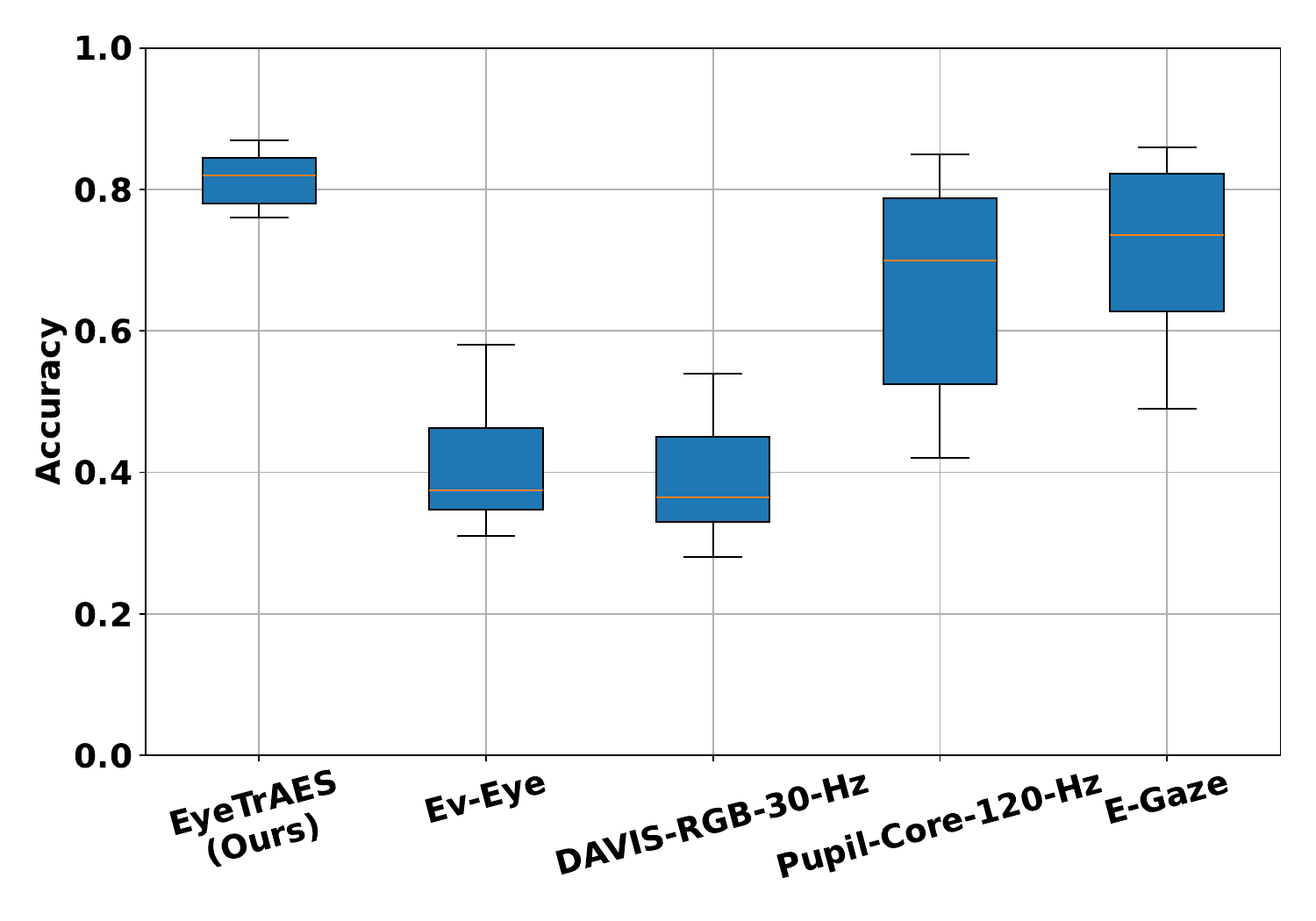}\label{fig:acc_diff_method_overall}
    }
    \subfigure[]{
    \includegraphics[width=0.45\textwidth]{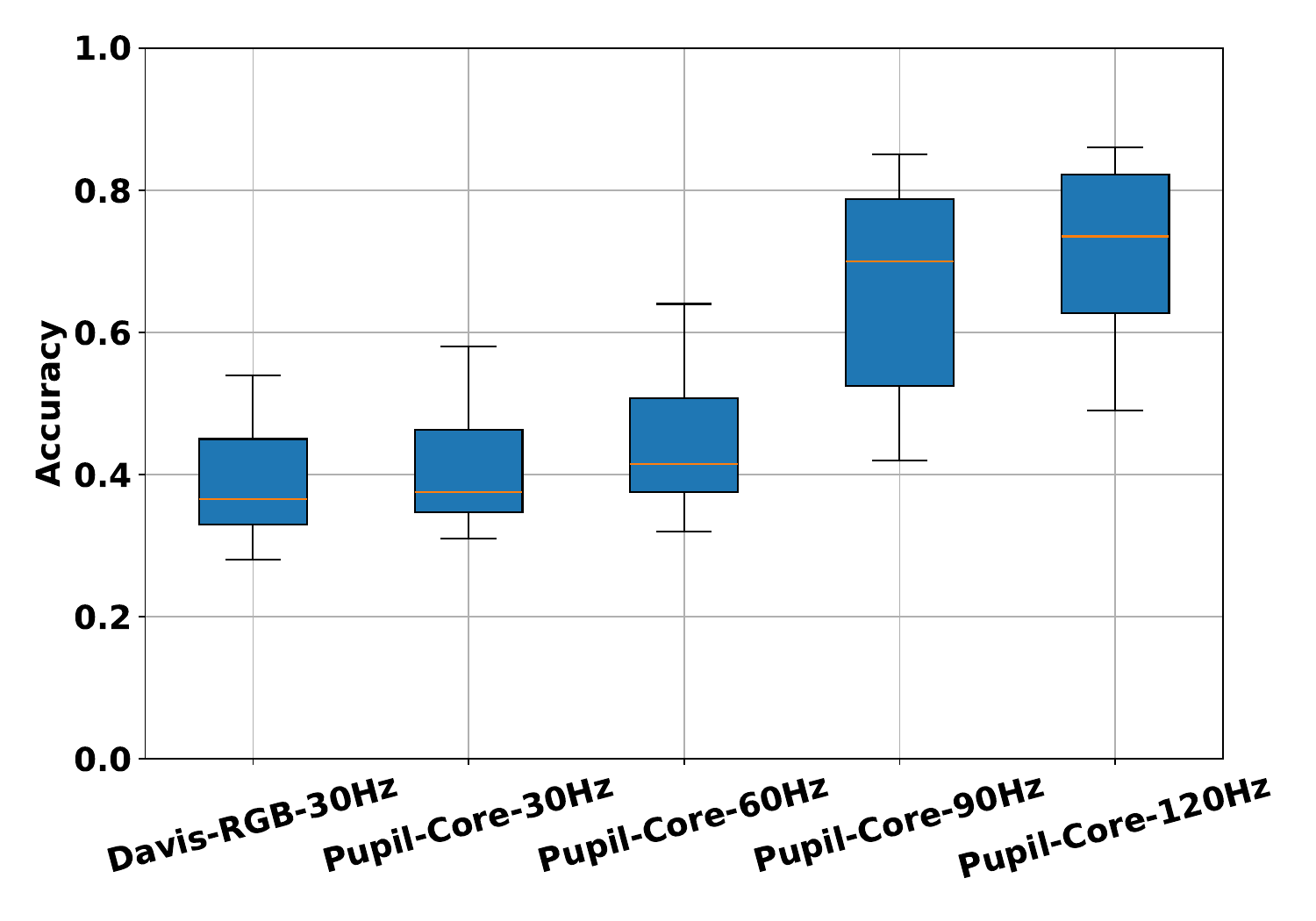}\label{fig:acc_diff_fps}
    }
    \subfigure[]{
    \includegraphics[width=0.45\textwidth]{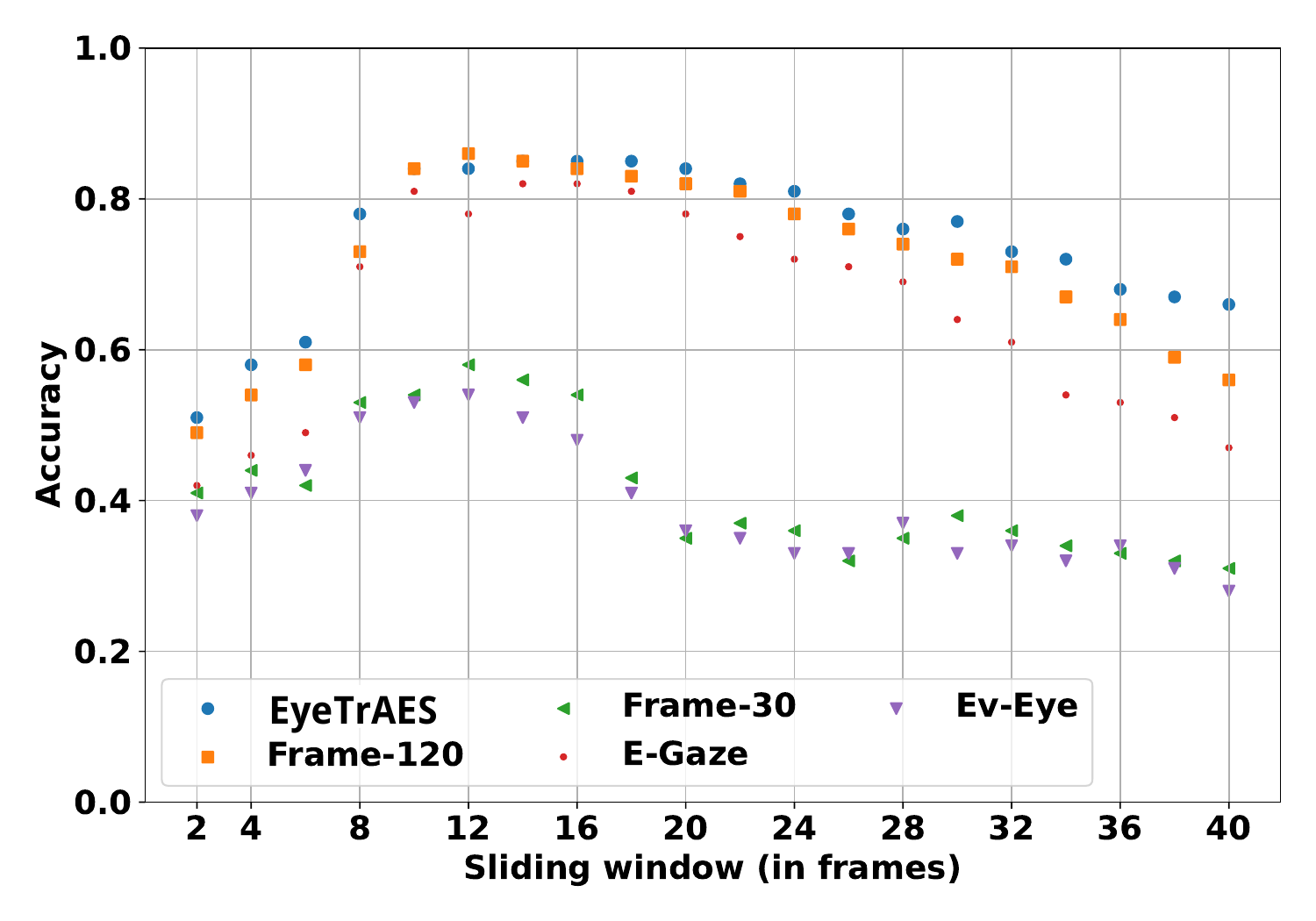}\label{fig:acc_diff_sliding}
    }
    \subfigure[]{
    \includegraphics[width=0.45\textwidth]{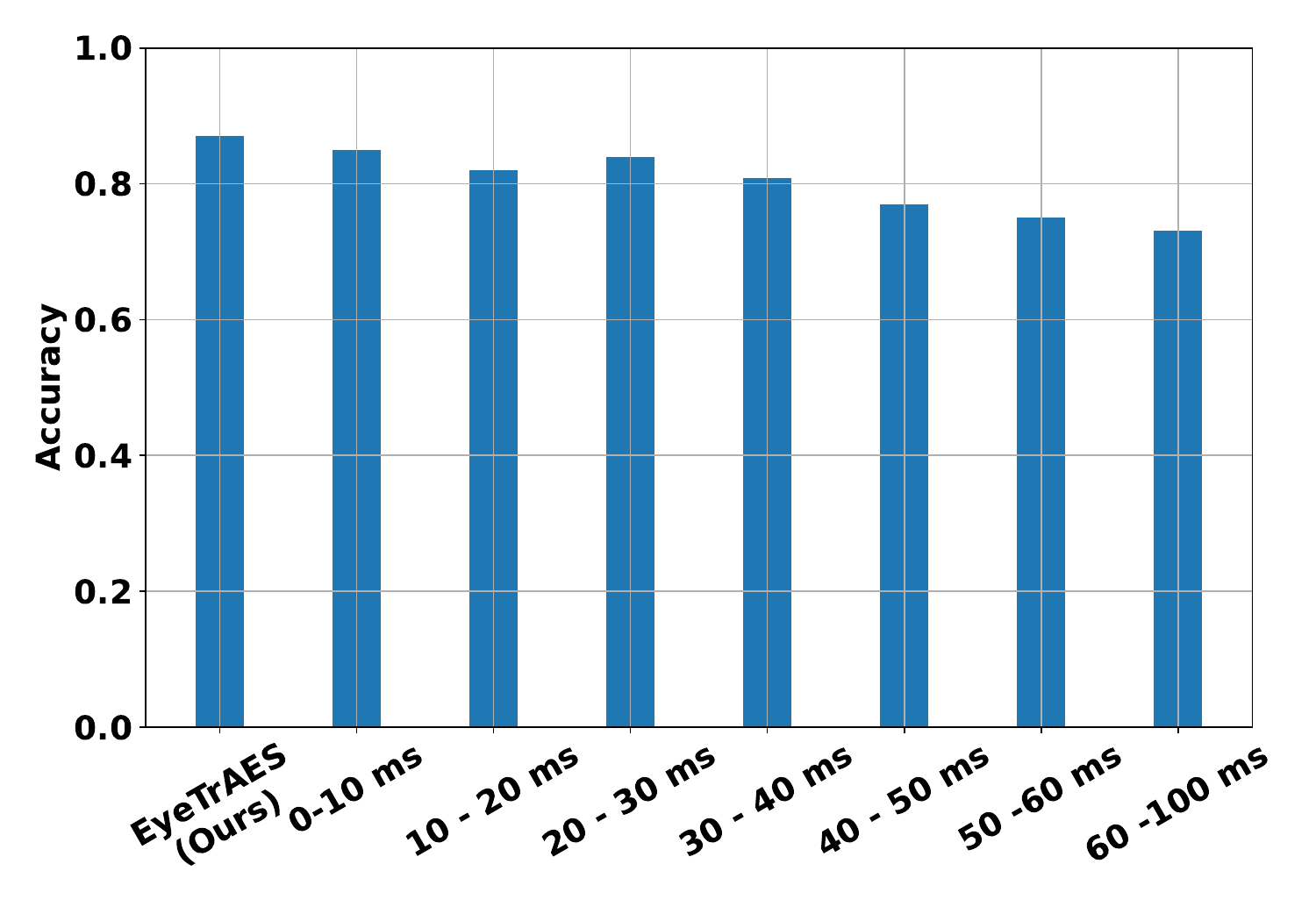}\label{fig:diff_slice_factor}
    }
    \caption{User authentication accuracy evaluated on \ourmethod dataset: (a) different baselines, (b) RGB frame-based approach at different FPS, (c) Average accuracy of \ourmethod{} for varying sliding window length, and (d) different slicing window regions from the adaptive slices.}
\end{figure}

As depicted in \figurename~\ref{fig:acc_diff_method_overall}, \ourmethod demonstrates higher accuracy than other alternatives,  achieving a median accuracy of $0.82$. Ev-Eye trained on our dataset demonstrated a significantly lower median accuracy of $0.43$. Additionally, the frame-based method with $30$ FPS exhibited an accuracy range of $0.34$ to $0.58$, with a median accuracy of $0.41$. However, as expected, the authentication accuracy for the RGB frame-based approach increases with increasing frame rate, with  $120$ FPS RGB streams resulting in a median accuracy of $0.7$. Higher frame rates can capture the rapid saccadic and micro-sacaddic eye movements with greater precision,  compared to a lower frame rate of 30 FPS. E-Gaze also demonstrated moderately good performance, with a median accuracy of $0.74$, attributed to its precise detection of the pupil region compared to other baseline approaches. Both Ev-Eye and DAVIS-RGB-30Hz utilize frame data captured at a lower rate=30 Hz, and thus have lower accuracy compared to the other baselines.

\changed{In addition to visualizing the performance differences between our proposed method and the baseline methods, we conducted a statistical analysis (using  a t-test) to compare the accuracies achieved by \ourmethod against those of the baseline methods. The t-test results as shown in Table~\ref{tab:ttest} demonstrate that the pupil tracking accuracy of \ourmethod is statistically significantly different than the accuracies achieved by the other alternatives, such as Ev-Eye, DAVIS-RGB-30Hz, Pupil-Core, and E-Gaze, with p-values less than $0.001$. These findings confirm that the accuracy gains of our approach are indeed statistically significant.}

\begin{table}[]
    \centering
    \caption{\changed{T-test Results Comparing Proposed Method and Baseline Methods}}
\label{tab:ttest}
    \begin{tabular}{|c|c|}
\hline
\textbf{Comparison}                 & \textbf{p-value} \\ \hline
\textbf{\ourmethod vs. Ev-Eye}         & $2.51 \times 10^{-15}$         \\ \hline
\textbf{\ourmethod vs. DAVIS-RGB-30Hz} & $6.09 \times 10^{-17}$         \\ \hline
\textbf{\ourmethod vs. Pupil-Core}     & $6.41 \times 10^{-11}$         \\ \hline
\textbf{\ourmethod vs. EGaze}          & $0.0000156$        \\ \hline
\end{tabular}
\end{table}

\subsubsection{Performance of frame-based approach under different FPS}
To understand the impact of different frame rates on user authentication accuracy, we use the Pupil-Core grayscale frames captured at 120 FPS and downsample them to \{90, 60, 30\} FPS by dropping the relevant intermediate frames. As demonstrated in \figurename~\ref{fig:acc_diff_fps} increasing the frame rate has a direct impact on the ability to estimate the kinematics of the eye movements, leading to higher accuracy. 

\subsubsection{Impact of sliding window length}

Having established the superiority of \ourmethod in extracting the pupil kinematics features even with the lower nominal frame rate, we next evaluate the impact of the length of the sliding window (over which the kinematic features are computed) on authentication accuracy. Figure~\ref{fig:acc_diff_sliding} depicts the average user authentication accuracy across varying sliding window length (measured in number of frames) of \ourmethod and baselines. The results indicate that the authentication accuracy generally improves with larger sliding window length, reaching a peak median accuracy of 0.87 for \ourmethod{} at a window length of 16 frames. However, beyond a certain point, increasing the window length leads to a decrease in accuracy as a larger window size leads to  over-accumulation of features, potentially representing different states of eye movement. A higher sliding window will also lead to a longer duration (i.e., lower responsiveness) for successful user authentication. Overall, a sliding window length of 10 seems to be a suitable choice, achieving classification accuracy of $82\%$ and providing a relatively low authentication response time. 


These findings suggest that the choice of sliding window length has a significant impact on the system's performance, and selecting an appropriate window length is crucial for achieving optimal accuracy.

\subsubsection{Impact of individual slicing factors}
We next study how our \ourmethodf adaptive slicing technique affects the authentication accuracy associated with different slice durations. We consider slice windows in 7 distinct ranges: 0–10, 10–20, 20–30, 30–40, 40–50, 50–60, and 60–100 ms. Because each authentication vector comprises 10 slices, possibly of varying duration, we first use \emph{dominant class labeling} to assign each authentication vector a specfic \emph{slice label}--i.e., the label corresponding to the modal slice duration. 
Figure~\ref{fig:diff_slice_factor} plots the average accuracy achieved by \ourmethod across different window ranges. We observe that the average accuracy is higher for samples corresponding to dominant slices $<40$ ms, with the accuracy progressively dropping slightly for samples with dominant slices $>60$ ms. More importantly, we see that the accuracy is relatively constant ($\approx$0.8$\pm$0.03) across all ranges, indicating that \ourmethod is reasonably successful in preserving salient eye movement features across different ranges.

\begin{figure}
    \centering
    \subfigure[]{
    \includegraphics[width=0.3\textwidth]{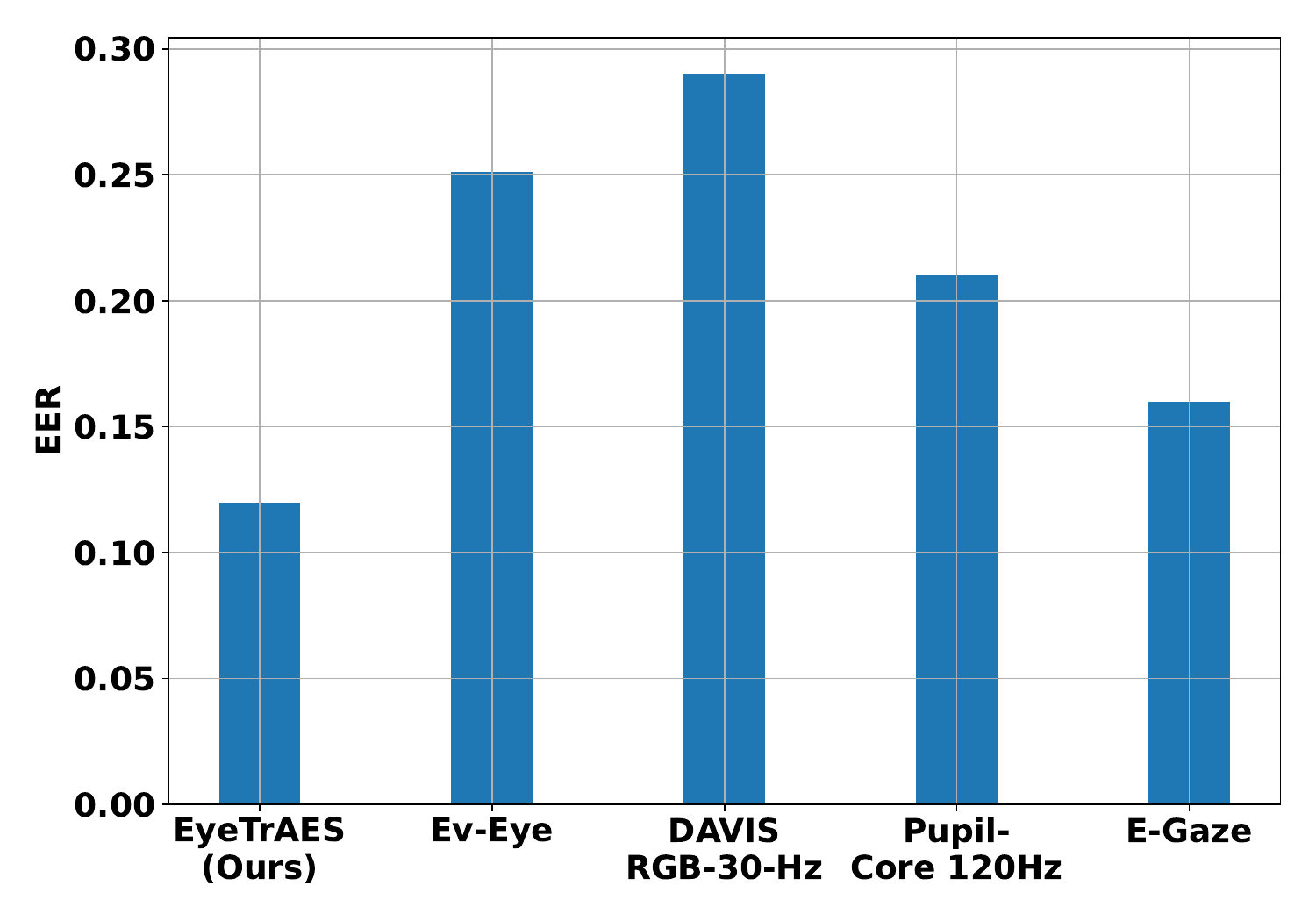}\label{fig:eer}
    }
    \subfigure[]{
    \includegraphics[width=0.3\textwidth]{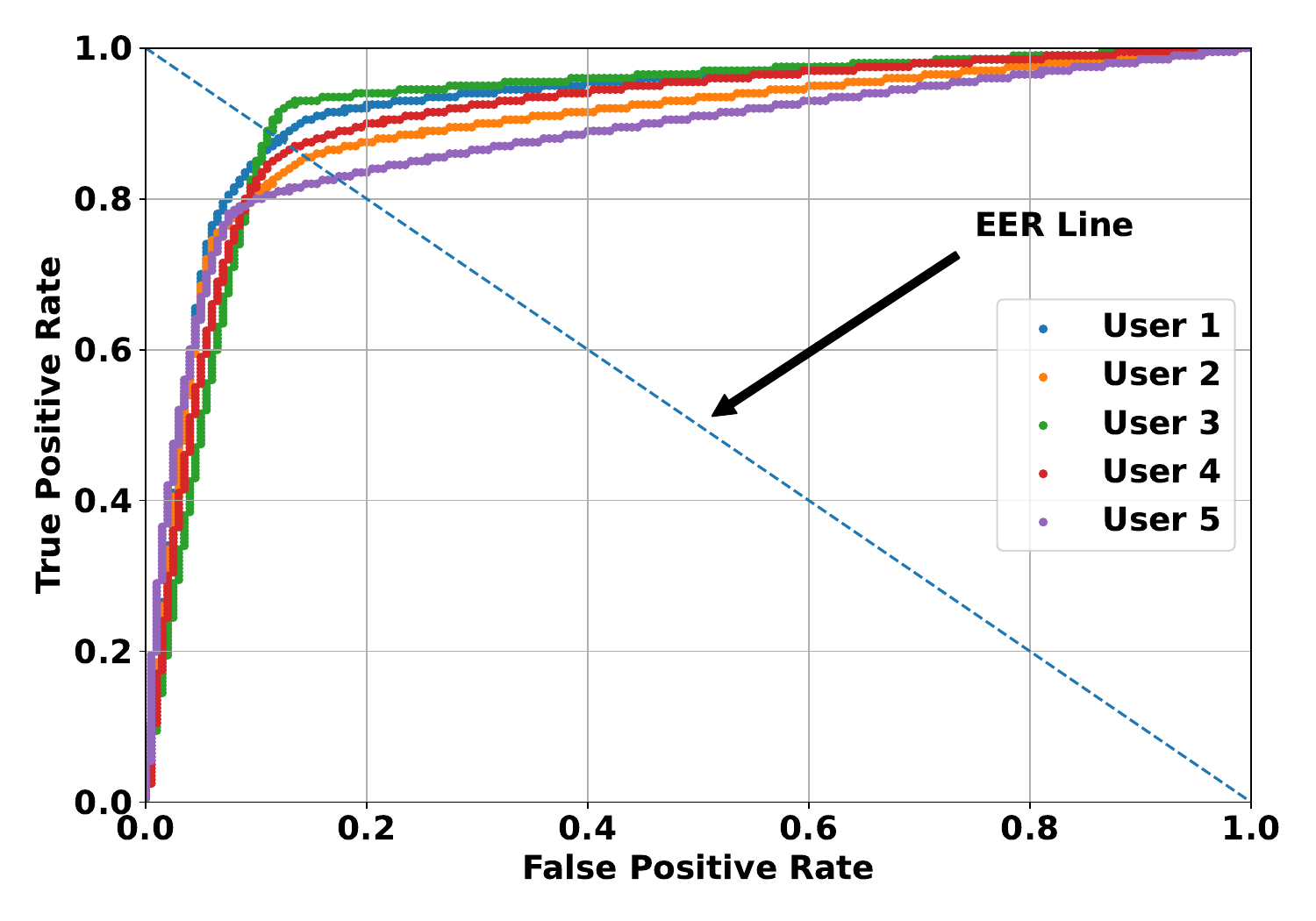}\label{fig:roc_user}
    }
    \subfigure[]{
    \includegraphics[width=0.3\textwidth]{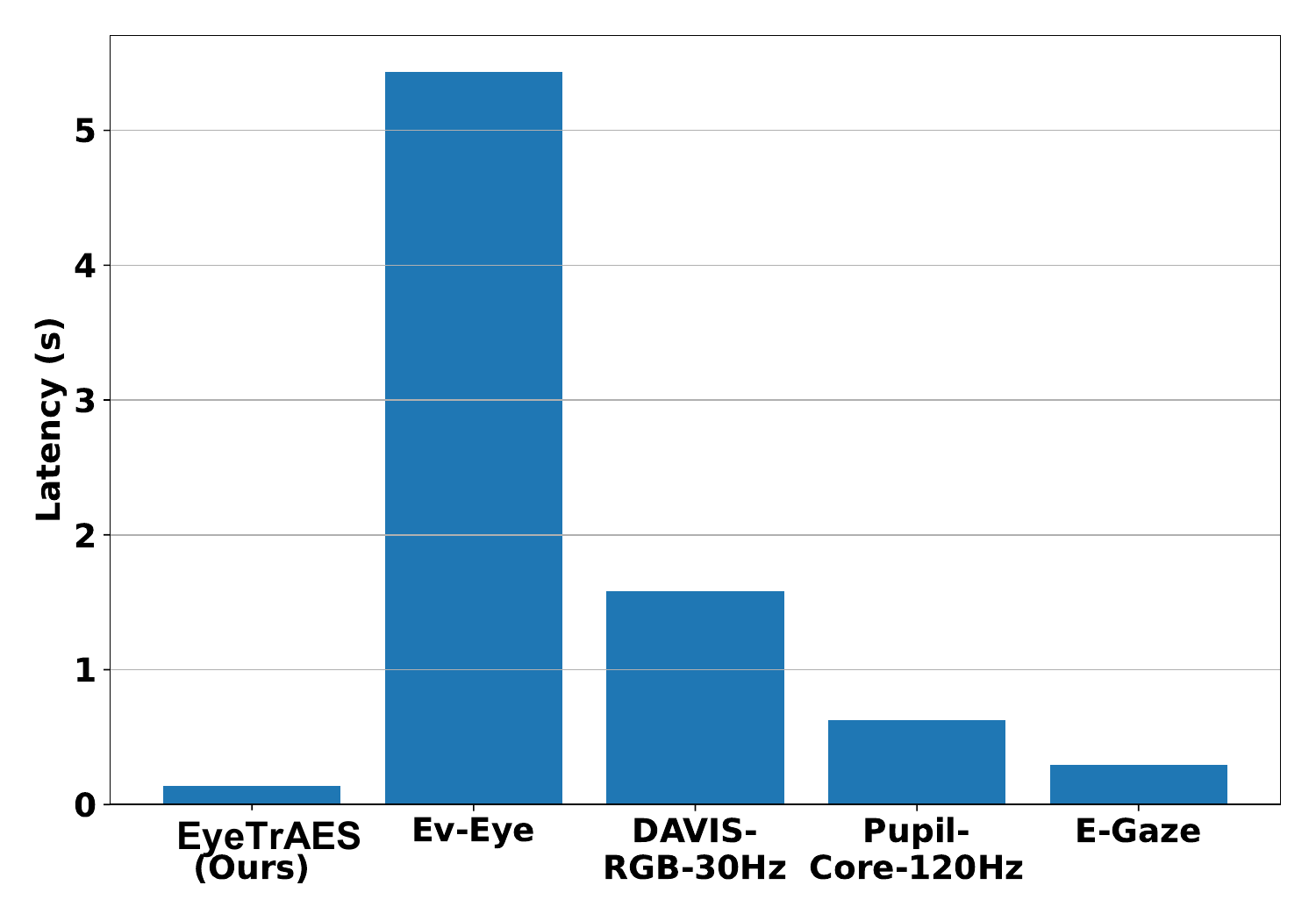}\label{fig:latency}
    }
    \caption{(a) Equal Error Rate (EER) for different approaches, (b) ROC curves across different users, (c) Response time in a valid authentication under each method.}
\end{figure}

\subsubsection{User re-authentication performance}
To verify a user's identity, we have also used the Equal Error Rate (EER), which is a commonly used metric used for biometric authentication systems. The EER represents the point on a Receiver Operating Characteristic (ROC) curve where the False Acceptance Rate (FAR) is equal to the False Rejection Rate (FRR). As shown in Figure~\ref{fig:eer}, compared to other baselines, \ourmethod{} performs better with a lower EER rate. A high EER indicates that the system is unable to effectively balance between false acceptances and false rejections. This can lead to authentication issues if impostors are frequently accepted or inconvenience if legitimate users are frequently rejected. Overall, \ourmethod{} shows an average EER of $0.12$, as also shown by the ROC curves in Figure~\ref{fig:roc_user} across 5 selected users from our dataset. 

\subsubsection{Authentication Responsiveness:} We have also calculated the authentication response time, which refers to the time required for the classifier to successfully authenticate a legitimate user for the first time, across all methods. This computation excludes the initial bootstrapping of the sliding window for generating features, assuming that the process has already been completed. Instead, we focus on the time taken from the computation of the latest pupil segmentation to the prediction time until a valid authentication is achieved. The response time for \ourmethod-based successful authentication is a maximum of $0.14$ sec.  As illustrated in Figure~\ref{fig:latency}, \ourmethod outperforms the baselines in terms of such response time. Ev-Eye exhibits the longest response time at $\approx 5.4$ seconds due to the inherent latency of its DNN-based pupil segmentation, while E-Gaze shows the shortest response time among the baselines at around $0.3$ seconds. The frame-based methods benefit from a higher frame rate (FPS) of 120, leading to faster accumulation of eye movement features and a response time of $0.62$ seconds. In contrast, the default $30$ FPS captured by DAVIS$346$ results in slower feature accumulation and a response time of $1.6$ seconds.

\begin{figure}
    \centering
    \subfigure[]{
    \includegraphics[width=0.3\textwidth]{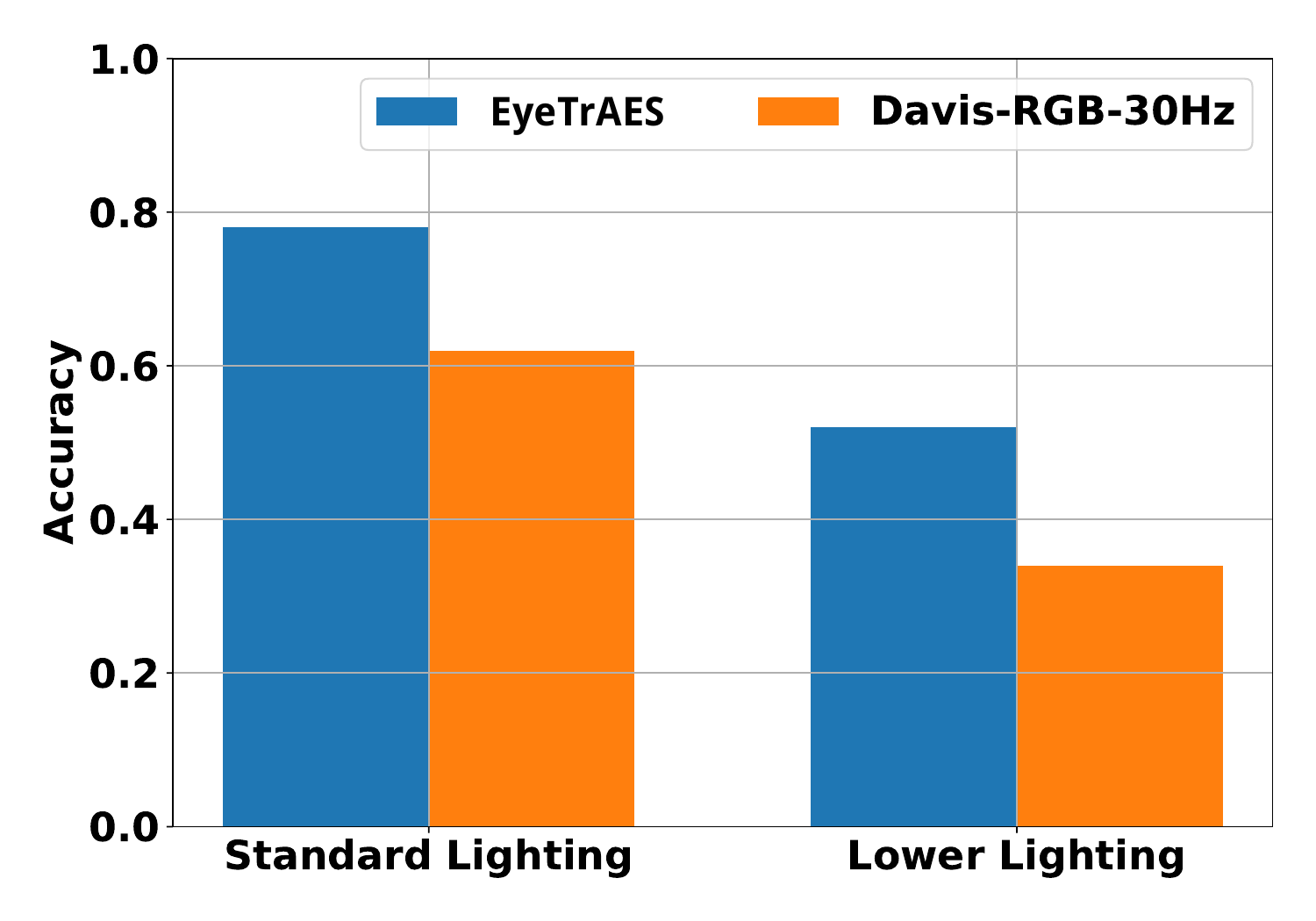}\label{fig:light}
    }
   \subfigure[]{
    \includegraphics[width=0.3\textwidth]{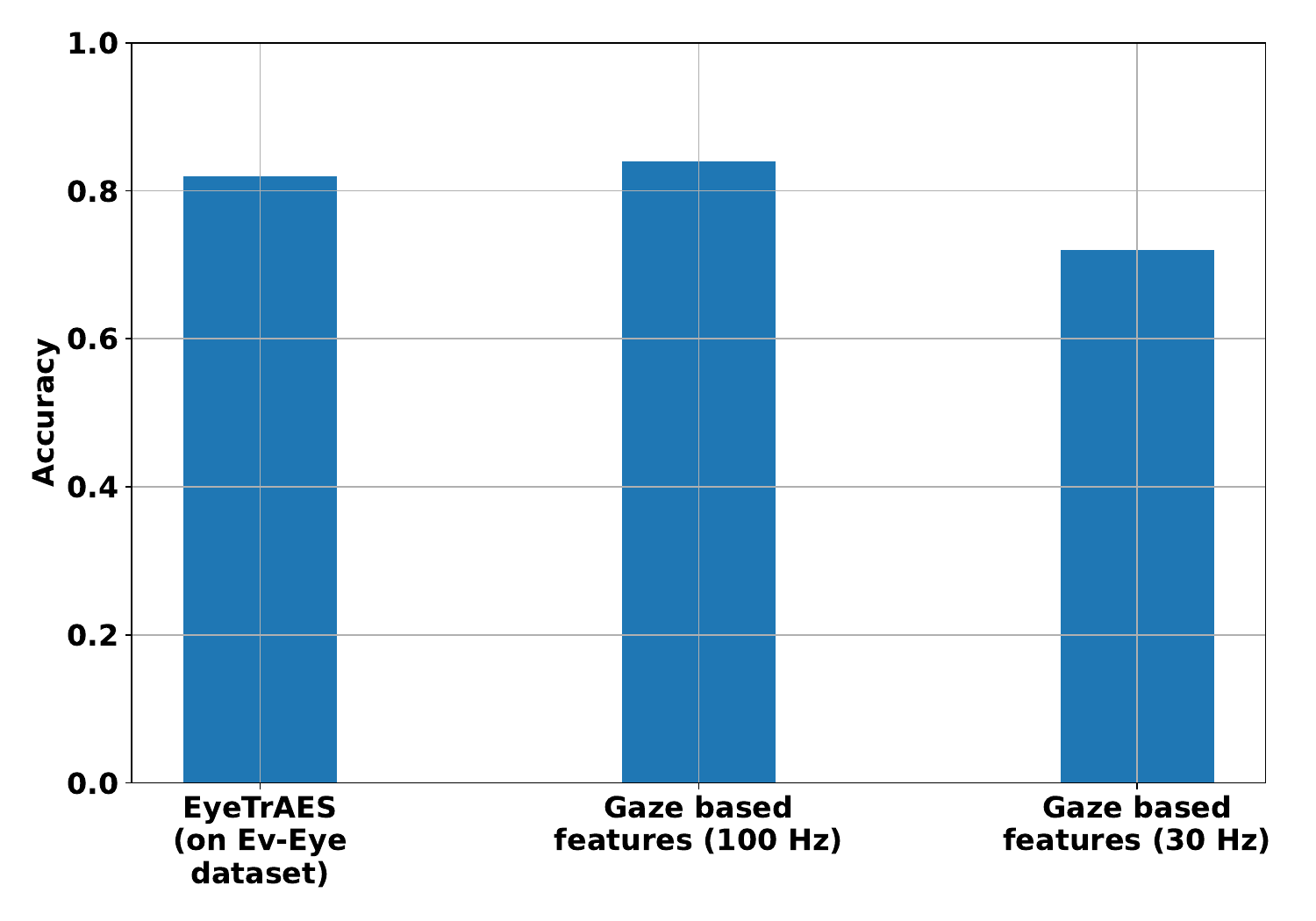}\label{fig:gaze}
    }
\subfigure[]{
    \includegraphics[width=0.3\textwidth]{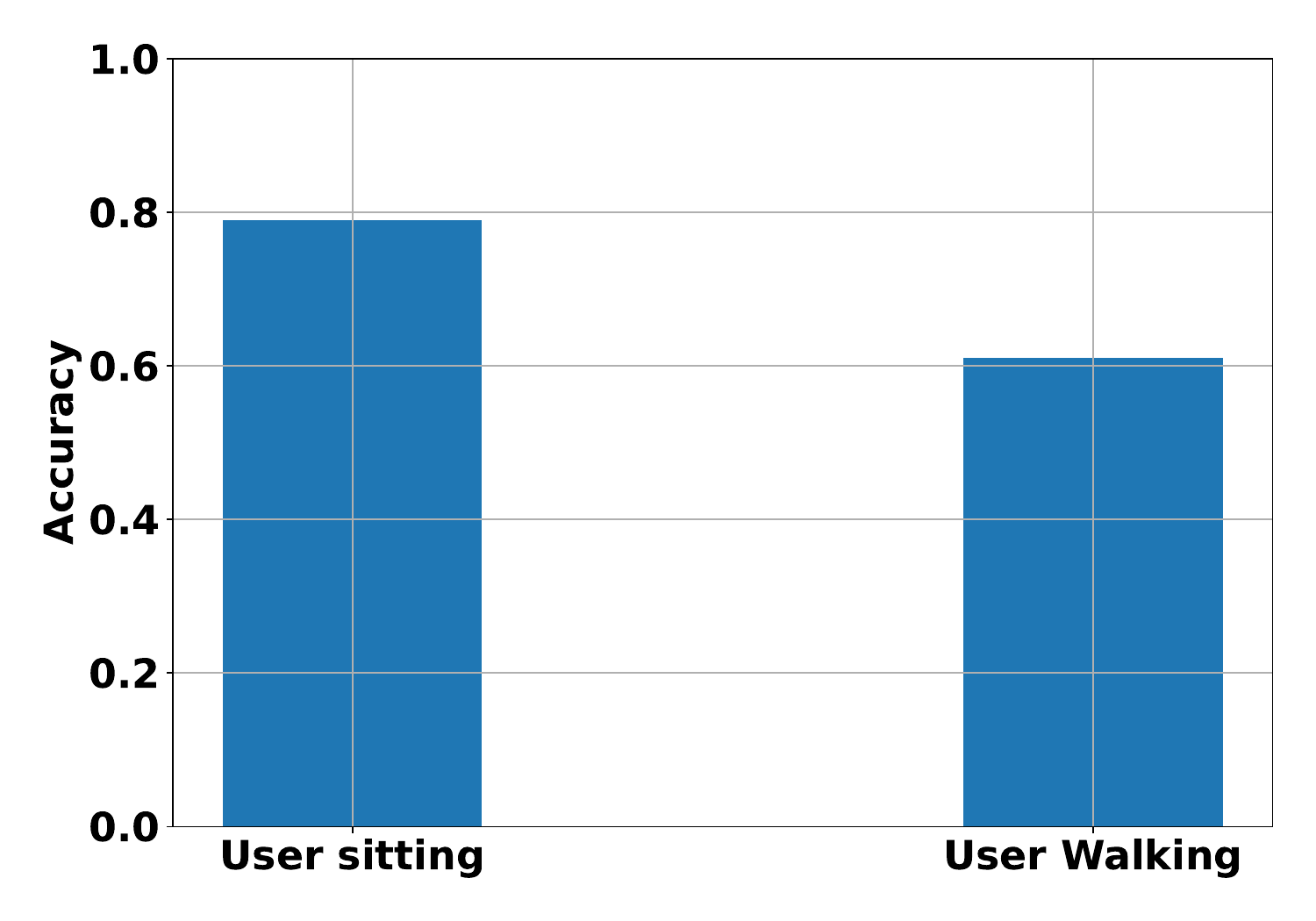}\label{fig:mot}
    }
    \caption{Mean Accuracy of (a) \ourmethod{} on Ev-Eye dataset, and with gaze-based features collected at 100 Hz and downsampled to 30 Hz, (b) under standard lighting conditions and at poor lighting conditions, (c) \ourmethod{} while the subject is sitting in the default setting vs when the subject is walking.}
\end{figure}

\subsubsection{Performance under different lighting conditions}
To evaluate the performance of \ourmethod{} under different lighting conditions, we explicitly collected additional data for a typical subject (id: 1) under poor lighting conditions. The light level is measured using a digital luminance light meter (LX1330B\footnote{\url{https://drmeter.com/products/lx1330b-digital-illuminance-light-meter}}). The measured light level is 24 Lux in the environment under poor lighting conditions, dropping to 8 Lux near the eye after wearing the custom headgear. We compare this subject authentication accuracy with the data collected in the default lighting condition, where the measured environmental and near-eye illuminance was 348 Lux and 65 Lux, respectively. Since Pupil-Core uses IR LEDs for eye tracking, its performance remains unaffected by lighting changes. We compared \ourmethod vs. the RGB-based baseline (DAVIS346-RGB-30Hz) under these different ambient lighting conditions. As shown in Figure~\ref{fig:light}, under poor lighting conditions, subject 1's authentication accuracy drops by $33\%$ to achieve accuracy $\sim$=62\%, as the rate of event generation also gets impacted under poor lighting conditions. However for DAVIS346-RGB-30Hz, the performance degradation is even more pronounced, with accuracy dropping by $\approx 45\%$ to $\sim$36\%. This drop occurs as the RGB method relies on segmenting out the pupil coordinates with color-based filtering techniques, which fails under poor lighting conditions.

\subsubsection{Performance from the Perspective of Usable Security}

\changed{In the context of biometric authentication, the evaluation of performance metrics such as False Acceptance Rate (FAR) and False Rejection Rate (FRR) is crucial for assessing both the security and usability aspects of the authentication system. Our proposed method demonstrates superior FAR performance compared to existing approaches, ensuring robust security. Specifically, our method achieved a FAR of $12\%$, which is significantly lower than the typical FAR values observed in similar systems, often ranging from $20\%$ to $34\%$. This reduction highlights the enhanced security capability of our method, making it well-suited, especially as a secondary, continuous authentication mechanism, for environments where security is a primary concern. Our method also achieved an FRR of $9\%$, which is lower compared to the FRR values of alternative systems ranging from $11\%$ to $32\%$. 
This balance between FAR and FRR is a critical factor in designing an effective biometric authentication system. More specifically, if used in a wearable device as a form of continuous, secondary authentication, having a lower FRR than FAR (as exhibited by \ourmethodnospace) is more critical. In addition, the annoying sporadic failure of legitimate authentication attempts can be further reduced by declaring an authentication failure only after multiple incorrect attempts: with a simple threshold of 3 consecutive failures, our effective FRR reduces to $\sim 7x10^{-4}$--i.e., roughly a failure rate of less than one in a thousand. }
\changed{Our method strikes an effective balance between these metrics, demonstrating superior performance in terms of both security and usability.}

\begin{table}[h]
\centering
\caption{\changed{Comparison of FAR and FRR Across Different Baselines}}
\label{tab:far_frr}
\begin{tabular}{|l|c|c|}
\hline
\textbf{Method} & \textbf{FAR (\%)} & \textbf{FRR (\%)} \\ \hline
\ourmethod (Ours) & 12 & 9 \\ \hline
Ev-Eye & 25 & 21 \\ \hline
DAVIS-RGB-30-Hz & 34 & 32 \\ \hline
PupilCore-120-Hz & 20 & 11 \\ \hline
E-Gaze & 24 & 19 \\ \hline
\end{tabular}
\end{table}

\subsection{Impact of Gaze vs Eye movement-related features in Authentication}
For biometric eye gaze-based authentication, previous works \cite{kasprowski2012first, george2016score} have proposed the Point of Gaze (PoG) coordinates, velocity, and acceleration of gaze changes as primary features. To understand how these extracted features perform on user authentication, we use the Ev-Eye dataset, where PoG coordinates are logged along with the event data using a Tobii Pro Glass 3\footnote{\url{https://www.tobii.com/products/eye-trackers/wearables/tobii-pro-glasses-3}}. From these gaze coordinates, we generate the features and pass them on to the classifier in two different settings: one where the gaze coordinates are passed at the default rate of collection, i.e., 100 Hz, and the other where the coordinates are downsampled to 30 Hz, emulating the sampling rate of an RGB frame or event accumulation window of 33 ms. From Figure~\ref{fig:gaze}, we observe that gaze-based classification, under the default sampling rate of 100 Hz, offers the highest accuracy ($\sim$84\%), as such high frequency data can better capture the accadic and micro-saccadic movements. Note, however, that gaze-based methods require concurrent sensing of both eyes, as well as additional knowledge of the viewing distance. However, if the gaze data is downsampled to 30 Hz (to mimic the sampling rate of a standard DAVIS346 RGB frame or event accumulation window of 33 ms), \ourmethod provides superior accuracy. \ourmethodf event-based pupil tracking is smoother and less noisy due to its use of a Kalman filter; in contrast, downsampled gaze coordinates have more discontinuity in the extracted gaze features, leading to slightly lower accuracy.




\section{Discussion}
\label{sec:discussion}

While our results demonstrate that \ourmethod provides significant enhancements to current capabilities pupil tracking and eye motion-based user authention, there are several open areas that require further investigation.

\noindent \emph{Single vs. Dual Eye Tracking}: \ourmethod currently uses wearable based near-eye tracking of only one eye, and thus cannot directly take advantage of gaze-related features such as saccades and fixation. We believe that our approach of tracking a single pupil’s movement suffices, as users typically tend to exhibit conjugate eye movement, moving both eyes in tandem. That said, it is possible that microscopic distinctions may exist between the pupillary movements of the right and left eyes, perhaps because of the differences in ocular muscle strength (most people have one dominant eye). There are two potential consequences of such possible distinctions. First, we may need to train separate \ourmethod models for each eye, as the pupillary movement of the dominant vs. non-dominant eye may exhibit salient differences. Second, it may be possible to further improve the user authentication accuracy by combining pupillary motion features from both eyes, as they may collectively encode subtler person-specific variations than available from tracking a single eye. Studying the implications of eye-specific variations, especially for users who may suffer from certain eye impairments such as myopia or amblyopia (aka lazy eye), is needed to demonstrate the applicability over a broad population.

\noindent \emph{Reliable Classification under Different Motion Conditions}: The \ourmethod dataset was collected in a controlled lab setting, with the user seated comfortably in a chair and gazing at the stimulus displayed on a stationary screen. While users were not restricted, unlike in EV-Eye, to keep their head stationary, we should note that our studies did not capture eye movement behavior under various real-world motion conditions, such as different ambulatory states (e.g., running, climbing) or different vehicular usage (e.g., buses, trains). There are two distinct reasons by which the captured pupillary motion during such real-world conditions may differ from those observed in the \ourmethod dataset. First, user movement can lead to continuous displacement of the wearable sensor relative to the human eye, leading in turn to noise in the captured event stream. This limitation is essentially due an imperfection in the sensing mechanism and can be overcome by simply ensuring a snug fit of the wearable device on the face. The second reason, however, is more fundamental: an individual’s \emph{pupil movement itself can be fundamentally altered due to such external context}—e.g., a user viewing a screen while walking may continuously glance, perhaps even without focusing their gaze, in multiple directions to maintain situational awareness. To perform a preliminary testing of this possibility, we collected additional pupil movement data, using our snugly-fitted wearable prototype, from a single subject while they were engaged in multiple activities such as walking on a treadmill or climbing stairs. As reported in Figure \ref{fig:mot}, we observed that \ourmethodf authentication accuracy for this user drops from 79\% (for test data collected while they are seated) to 61\% (for test data collected during such activities). These preliminary results suggest that \ourmethodf accuracy can be possibly enhanced by incorporating additional macro-motion features (e.g., captured by smartglass-mounted inertial sensors) into the authentication classifier.


\noindent \emph{Native SNN-based Processing of Events}: As explained earlier, we adopt a strategy of frame-based event accumulation instead of processing the events natively using an SNN model, simply because of the current lack of embedded neuromorphic processors that can support efficient SNN execution. We adopted this decision because preliminary studies indicating that a software-based emulation of an SNN, using the SpikingJelly framework~\cite{fang2023spikingjelly}, is simply too slow and can process at most 9-10 ``frames” per second, even on a powerful Jetson ORIN platform. Should neuromorphic processors become available, we anticipate that SNN-based approaches will prove to be more efficient, at least until the pupil segmentation and tracking stage. The resulting change in the frequency and accuracy of the stream of inferred pupil location data is likely to require a modification of the set of pupillary kinematic features and the corresponding Random Forest classifier model. This remains future work.

\section{Conclusion}\label{sec:conclusion}
\changed{In this paper, we have presented a novel approach for fine-grained low-latency pupillary movement tracking using event cameras. Our approach, \ourmethod, uses a novel adaptive event accumulation technique coupled with a light-weight pupil segmentation algorithm to track the eye pupil region with significantly higher accuracy and lower latency -- pupil segmentation IoU score $\sim$=92\% while incurring frame processing latency of only $\sim$4.7 ms. Further, as an illustrative application, we showcase the extracted microscopic eye kinematic features (such as pupil velocity and acceleration) from the high-fidelity pupil tracking data exhibit distinctive trends across users and can in turn be used as a means for robust user authentication. Our exemplar application, \ourmethodnospace-based user authentication, has several key advantages. Firstly, it offers high accuracy and reliability, as eye movements are unique to each individual and can serve as reliable biometric identifiers. Secondly, it provides a seamless and non-intrusive authentication experience, as users can be authenticated simply by looking at a screen or a camera. Thirdly, it is more robust to variations in lighting conditions and facial expressions, making it suitable for real-world applications. We have demonstrated the effectiveness of our approach through experiments and evaluations, showing that it outperforms traditional RGB camera-based authentication systems by achieving median user authentication accuracy $\sim$=0.82 and lower latency of $\sim$=12ms, showing its ability to achieve real-time on-device execution.}

\bibliographystyle{ACM-Reference-Format}
\bibliography{main}


\end{document}